\documentclass[10pt,twocolumn,letterpaper]{article}

\usepackage{cvpr}
\usepackage{times}
\usepackage{epsfig}
\usepackage{graphicx}
\usepackage{amsmath}
\usepackage{amssymb}

\usepackage{array}
\usepackage{multirow}
\usepackage{longtable}
\usepackage{tabularx}
\usepackage{subfigure}
\usepackage{algorithm}
\usepackage{algorithmicx}
\usepackage{algpseudocode}
\usepackage{arydshln}
\usepackage{pifont}
 \usepackage{url}

\newcommand{\cmark}{\ding{51}}%
\newcommand{\xmark}{\ding{55}}%
\algnewcommand\algorithmicinput{\textbf{Input:}}
\algnewcommand\INPUT{\item[\algorithmicinput]}
\algnewcommand\algorithmicoutput{\textbf{Output:}}
\algnewcommand\OUTPUT{\item[\algorithmicoutput]}
\def\BB{\textcolor{blue}}
\def\RR{\textcolor{black}}

\DeclareMathOperator*{\argmin}{arg\,min}
\global\long\def\defeq{\stackrel{\mathrm{{\scriptscriptstyle def}}}{=}}

\global\long\def\otilde{\widetilde{O}}
\usepackage[pagebackref=true,breaklinks=true,letterpaper=true,colorlinks,bookmarks=false]{hyperref}

\cvprfinalcopy 

\def\cvprPaperID{2677} 

\ifcvprfinal\fi
\def\BB{\textcolor{black}}
\begin{document}

\title{Filter Pruning via Geometric Median \\ for Deep Convolutional Neural Networks Acceleration}

\author{
Yang He\\
CAI, University of Technology Sydney\\
{\tt\small yang.he-1@student.uts.edu.au}
\and
Ping Liu \\
JD.com \\
{\tt\small pino.pingliu@gmail.com}
\and
Ziwei Wang\\
Information Science Academy, CETC\\
{\tt\small wangziwei26@gmail.com}
\and
Yi Yang\\
CAI, University of Technology Sydney\\
{\tt\small yee.i.yang@gmail.com}
}

\author{Yang He$^{1}$ \hspace{2em} Ping Liu$^{1,2}$ \hspace{2em} Ziwei Wang$^{3}$ \hspace{2em} Zhilan Hu$^{4}$ \hspace{2em} Yi Yang$^{1,5}\thanks{Corresponding Author. Part of this work was done when Yi Yang was visiting Baidu Research during his Professional Experience Program.}$\\
$^{1}$CAI, University of Technology Sydney \hspace{2em}
$^{2}$JD.com\\
$^{3}$Information Science Academy, CETC  \hspace{2em}
$^{4}$Huawei  \hspace{2em}
$^{5}$Baidu Research \\
{\tt\small \{yang.he-1\}@student.uts.edu.au~~\{pino.pingliu,wangziwei26,yee.i.yang\}@gmail.com}
}

\maketitle

\begin{abstract}

Previous works utilized ``smaller-norm-less-important" criterion to prune filters with smaller norm values in a convolutional neural network. In this paper, we analyze this norm-based criterion and point out that its effectiveness depends on two requirements that are not always met: (1) the norm deviation of the filters should be large; (2) the minimum norm of the filters should be small. To solve this problem, we propose a novel filter pruning method, namely Filter Pruning via Geometric Median (FPGM), to compress the model regardless of those two requirements. Unlike previous methods, FPGM compresses CNN models by pruning filters with redundancy, rather than those with ``relatively less" importance. When applied to two image classification benchmarks, our method validates its usefulness and strengths.
Notably, on \BB{CIFAR-10}, FPGM reduces more than 52\% FLOPs on ResNet-110 with even 2.69\% relative accuracy improvement. Moreover, on ILSVRC-2012, FPGM reduces more than 42\% FLOPs on ResNet-101 without top-5 accuracy drop, which has advanced the state-of-the-art.
Code is publicly available on GitHub: https://github.com/he-y/filter-pruning-geometric-median
\end{abstract}

\section{Introduction}\label{section:Introduction}

\begin{figure}[!ht]
\begin{centering}
\includegraphics[width=0.47\textwidth]{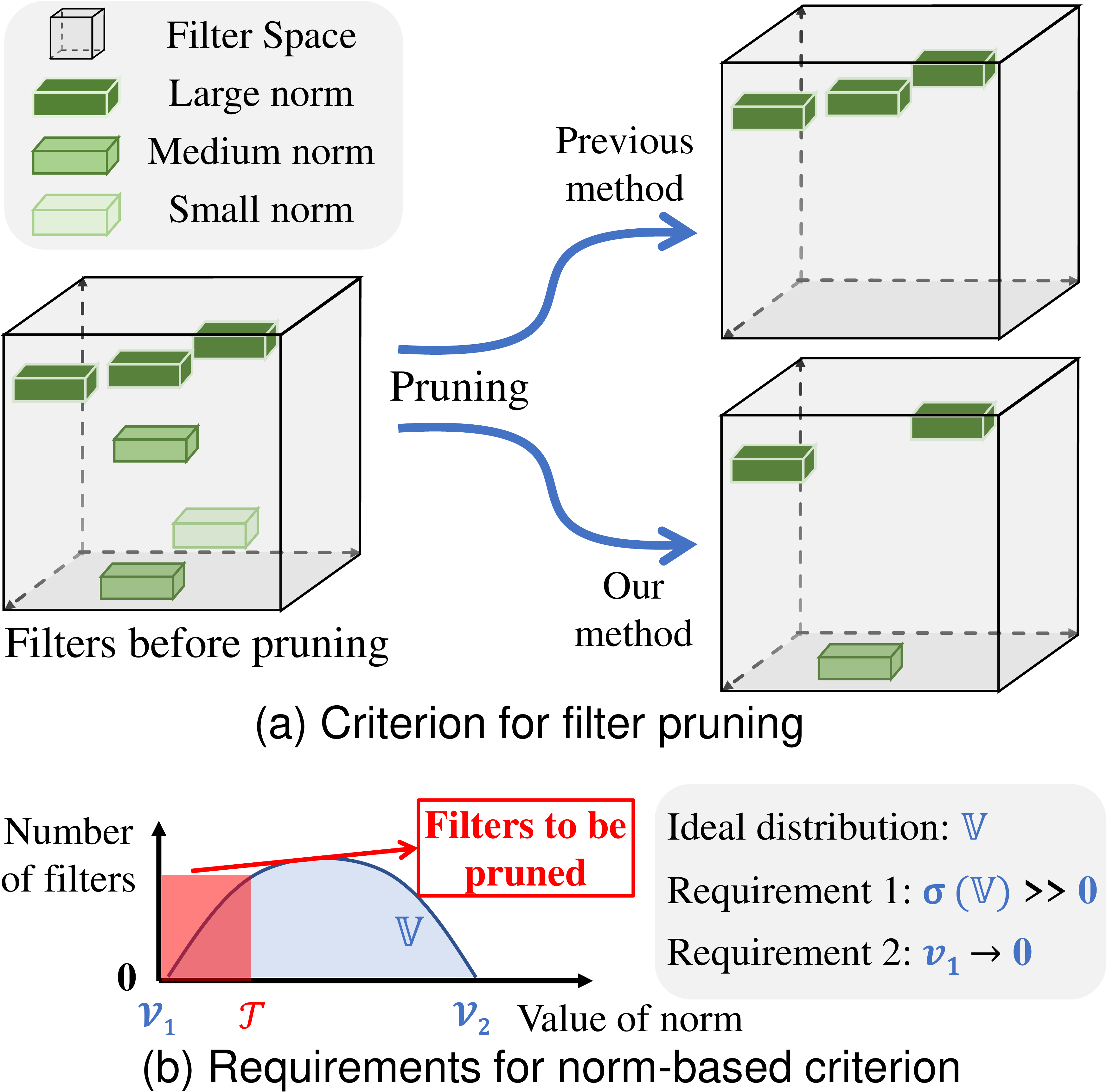} 
\par\end{centering}
\caption{
An illustration of (a) the pruning criterion for norm-based approach and the proposed method; (b) requirements for norm-based filter pruning criterion. In (a), the green boxes denote the filters of the network, where deeper color denotes larger norm of the filter. For the norm-based criterion, only the filters with the largest norm are kept based on the assumption that smaller-norm filters are less important. In contrast, the proposed method prunes the filters with redundant information in the network. In this way, filters with different norms indicated by different intensities of green may be retained. In (b), the blue curve represents the ideal norm distribution of the network, and the $v_1$ and $v_2$ is the minimal and maximum value of norm distribution, respectively. To choose the appropriate threshold $\mathcal{T}$ (the red shadow), two requirements should be achieved, that is, the norm deviation should be large, and the minimum of the norm should be arbitrarily small.
}
\label{fig:assumption} 
\end{figure}
The deeper and wider architectures of deep CNNs bring about the superior performance of computer vision tasks~\cite{dong2019search,Yawei2019Taking,zhu2019simreal}. However, they also cause the prohibitively expensive computational cost and make the model deployment on mobile devices hard if not impossible. Even the latest architecture with high efficiencies, such as residual connection~\cite{he2016deep} or inception module~\cite{szegedy2015going}, has millions of parameters requiring billions of float point operations (FLOPs)~\cite{he2018soft}.
Therefore, it is necessary to attain the deep CNN models which have relatively low computational cost but high accuracy.

Recent developments on pruning can be divided into two categories, \textit{i.e.}, weight pruning~\cite{han2015learning,carreira2018learning} and filter pruning~\cite{li2016pruning,yu2018nisp}.
Weight pruning directly deletes weight values \BB{in a filter} which may cause unstructured sparsities. This irregular structure makes it difficult to leverage the high-efficiency Basic Linear Algebra Subprograms (BLAS) libraries~\cite{Luo_2017_ICCV}.
In contrast, filter pruning directly discards \BB{the whole selected filters} and leaves a model with regular structures. Therefore, filter pruning is more preferred for accelerating the networks and decreasing the model size.

Current practice~\cite{li2016pruning,ye2018rethinking,he2018soft} performs filter pruning by following the ``smaller-norm-less-important" criterion, which believes that filters with smaller norms can be pruned safely due to their less importance. As shown in the top right of Figure~\ref{fig:assumption}(a), after calculating norms of filters in a model, a pre-specified threshold $\mathcal{T}$ is utilized to select filters whose norms are smaller than it. 


However, as illustrated in  Figure~\ref{fig:assumption}(b), there are two prerequisites to utilize this ``smaller-norm-less-important" criterion. First, the deviation of filter norms \BB{should be significant.} This requirement makes the searching space for threshold $\mathcal{T}$ wide enough \BB{so that separating those filters needed to be pruned would be an easy task.} Second, the norms of those filters which can be pruned should be arbitrarily small, \textit{i.e.}, close to zero; in other words, the filters with smaller norms are expected to make absolutely small contributions, rather than relatively less but positively large contributions, to the network. An ideal norm distribution when satisfactorily meeting those two requirements is illustrated as the blue curve in Figure~\ref{fig:assumption}. Unfortunately, based on our analysis and experimental observations, this is not always true.


\BB{To address the problems mentioned above, we propose a novel filter pruning approach, named Filter Pruning via Geometric Median (FPGM). 
Different from the previous methods which prune filters with \textbf{\emph{relatively less contribution}}, FPGM chooses the filters with \textbf{\emph{the most replaceable contribution}}.
Specifically, we calculate the Geometric Median (GM)~\cite{fletcher2008robust} of the filters within the same layer. According to the characteristics of GM, the filter(s) $\mathcal{F}$ near it can be represented by the remaining ones. Therefore, pruning those filters will not have substantial negative influences on model performance. Note that FPGM does not utilize norm based criterion to select filters to prune, which means its performance will not deteriorate even when failing to meet requirements for norm-based criterion.}





\textbf{Contributions.} We have three contributions:

(1)~We analyze the norm-based criterion utilized in previous works, which prunes the relatively less important filters. We elaborate on its two underlying requirements which lead to its limitations;

(2)~We propose FPGM to prune the most replaceable filters containing redundant information, which can still achieve good performances when norm-based criterion fails;

(3)~The extensive experiment on two benchmarks demonstrates the effectiveness and efficiency of FPGM. 

\section{Related Works}
Most previous works on accelerating CNNs can be roughly divided into four
categories, namely, \emph{matrix decomposition}~\cite{zhang2016accelerating,tai2015convolutional}, \emph{low-precision
weights}~\cite{zhu2016trained,zhou2017incremental,Son_2018_ECCV}, knowledge distilling~\cite{hinton2015distilling,kim2018paraphrasing} and \emph{pruning}.
\emph{Pruning}-based approaches aim to remove the unnecessary connections of the neural network \cite{han2015learning,li2016pruning,liu2018frequency}.
Essentially, \textbf{\emph{weight pruning}} always results in unstructured models, which makes it hard to deploy the efficient BLAS library, while \textbf{\emph{filter pruning}} not only reduces the storage usage on devices but also decreases computation cost to accelerate the inference. We could roughly divide the filter pruning methods into two categories by whether the training data is utilized to determine the pruned filters, that is, \emph{data dependent} and \emph{data independent} filter pruning. Data independent method is more efficient than data dependent method as the \BB{utilizing} of training data is computation consuming.

\textbf{Weight Pruning.}
Many recent works~\cite{han2015learning,han2015deep,guo2016dynamic,tung2018clip,carreira2018learning,he2018soft,zhang2018systematic,dong2017learning} focus on pruning fine-grained weight of filters.
For example, \cite{han2015learning} proposes an iterative method to discard the small weights whose values are below the predefined threshold.
~\cite{carreira2018learning} formulates pruning as an optimization problem of finding the weights that minimize the loss while satisfying a pruning cost condition.

\textbf{Data Dependent Filter Pruning.}
Some filter pruning approaches~\cite{Liu_2017_ICCV,Luo_2017_ICCV,He_2017_ICCV,molchanov2016pruning,dubey2018coreset,suau2018principal,yu2018nisp,wang2018exploring,zhuang2018discrimination,he2018adc,huang2018learning,lin2019towards} need to utilize training data to determine the pruned filters.
\cite{Luo_2017_ICCV} adopts the statistics information from the next layer to guide the filter selections.
\cite{dubey2018coreset} aims to obtain a decomposition by minimizing the reconstruction error of training set sample activation.
\cite{suau2018principal} proposes an inherently data-driven method which use  Principal Component Analysis (PCA) to specify the proportion of the energy that should be preserved.
\cite{wang2018exploring} applies subspace clustering to feature maps to eliminate the redundancy in convolutional filters.

\textbf{Data Independent Filter Pruning.}
Concurrently with our work, some data independent filter pruning strategies~\cite{li2016pruning,he2018soft,ye2018rethinking,zhuo2018scsp} have been explored. 
\cite{li2016pruning} utilizes an $\ell_{1}$-norm criterion to prune unimportant filters.
\cite{he2018soft} proposes to select filters with a $\ell_{2}$-norm criterion and prune those selected filters in a soft manner.
\cite{ye2018rethinking} proposes to prune models by enforcing sparsity on the scaling parameter of batch normalization layers. 
\cite{zhuo2018scsp} uses spectral clustering on filters to select unimportant ones.

\textbf{Discussion.}
To the best of our knowledge, only one previous work reconsiders the smaller-norm-less-important criterion~\cite{ye2018rethinking}.
We would like to highlight our advantages compared to this approach as below:
(1) \cite{ye2018rethinking} pays more attention to enforcing sparsity on the scaling parameter in the batch normalization operator, which is not friendly to the structure without batch normalization. On the contrary, our approach is not limited by this constraint. 
(2) After pruning channels selected, \cite{ye2018rethinking} need fine-tuning to reduce performance degradation. However, our method combines the pruning operation with normal training procedure. Thus extra fine-tuning is not necessary.
(3) Calculation of the gradient of scaling factor is needed for \cite{ye2018rethinking}; therefore lots of computation cost are inevitable, whereas our approach could accelerate the neural network \BB{without calculating the gradient of scaling factor.}

\section{Methodology}

\subsection{Preliminaries}\label{Preliminary}

We formally introduce symbols and notations in this subsection. We assume that a neural network has $L$ layers.
We use $N_{i}$ and $N_{i+1}$, to represent the number of input channels and the output channels for the $i_{th}$ convolution layer, respectively. $\mathcal{F}_{i,j}$ represents the $j_{th}$ filter of the $i_{th}$ layer, then the dimension of filter $\mathcal{F}_{i,j}$ is $\mathbb{R}^{N_{i}\times K\times K}$, where $K$ is the kernel size of the network\footnote{Fully-connected layers equal to convolutional layers with $k=1$}.
The $i_{th}$ layer of the network $\mathbf{W}^{(i)}$ could be represented by $\{\mathcal{F}_{i,j}, 1 \leq j \leq N_{i+1}\}$. The tensor of connection of the deep CNN network could be parameterized by $\{\mathbf{W}^{(i)}\in\mathbb{R}^{N_{i+1}\times N_{i}\times K\times K},1\leq i\leq L\}$.

\subsection{Analysis of Norm-based Criterion}
\label{section:norm_distribution}

\begin{figure}[h!]
\begin{centering}
\includegraphics[width=0.47\textwidth]{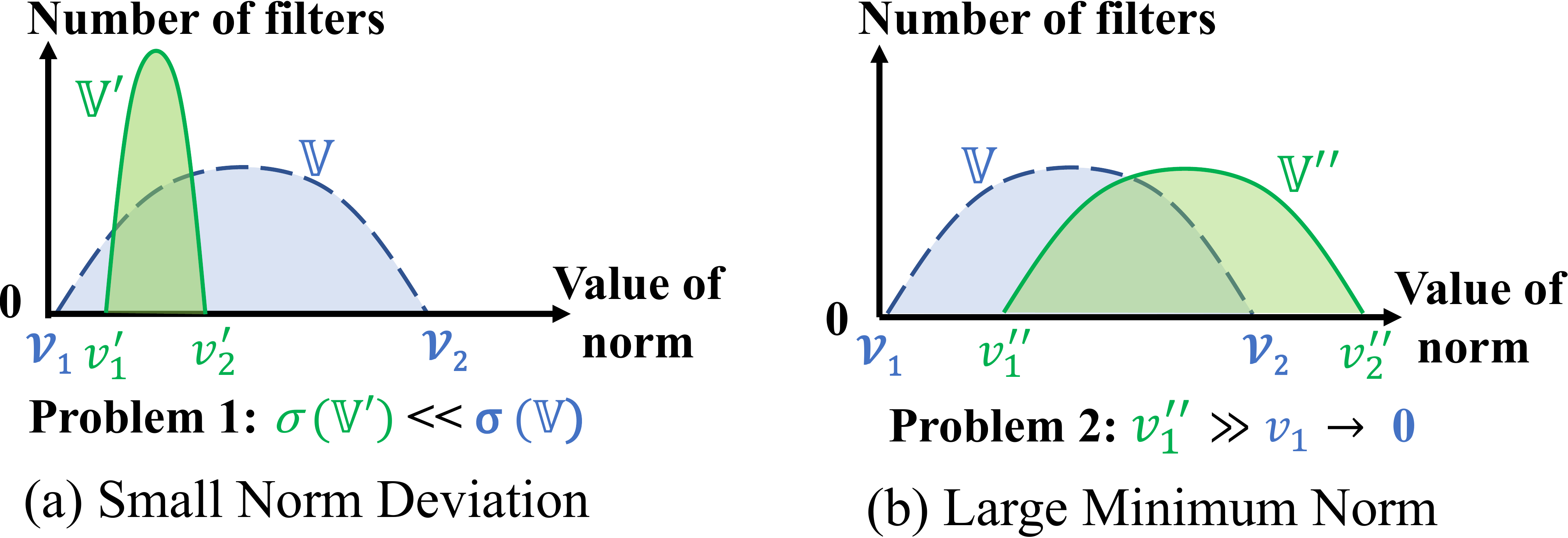} 
\par\end{centering}
\caption{
\BB{Ideal and Reality of the norm-based criterion:} (a) Small Norm Deviation and (b) Large Minimum Norm. The blue dashed curve indicates the ideal norm distribution, and the green solid curve denotes the norm distribution might occur in real cases.
}
\label{fig:problem} 
\end{figure}

Figure~\ref{fig:assumption} gives an illustration for the two requirements for successful utilization of the norm-based criterion. \BB{However, these requirements may not always hold, and it might lead to unexpected results.} The details are illustrated in Figure~\ref{fig:problem}, in which the blue dashed curve and the green solid curve indicates the norm distribution in ideal \BB{and real cases,} respectively.

(1) \emph{Small Norm Deviation}. The deviation of filter norm distributions might be too small, which means the norm values are concentrated to a small interval, as shown in Figure~\ref{fig:problem}(a). A small norm deviation leads to a small search space, which makes it difficult to find an appropriate threshold to select filters to prune.

(2) \emph{Large Minimum Norm}. The filters with the minimum norm may not be arbitrarily small, as shown in the Figure~\ref{fig:problem}(b), $v_1''>>v_1 \rightarrow 0$. Under this condition, those filters considered as the least important still contribute significantly to the network, which \BB{means} every filter is highly informative. Therefore, pruning those filters with minimum norm values will cast a negative effect on the network.


\subsection{Norm Statistics in Real Scenarios}
\begin{figure*}[t!]
\center
\subfigure[ResNet-110 (linear x-scale)]{
\label{fig:norm_dist1}
\includegraphics[width=0.235\linewidth]{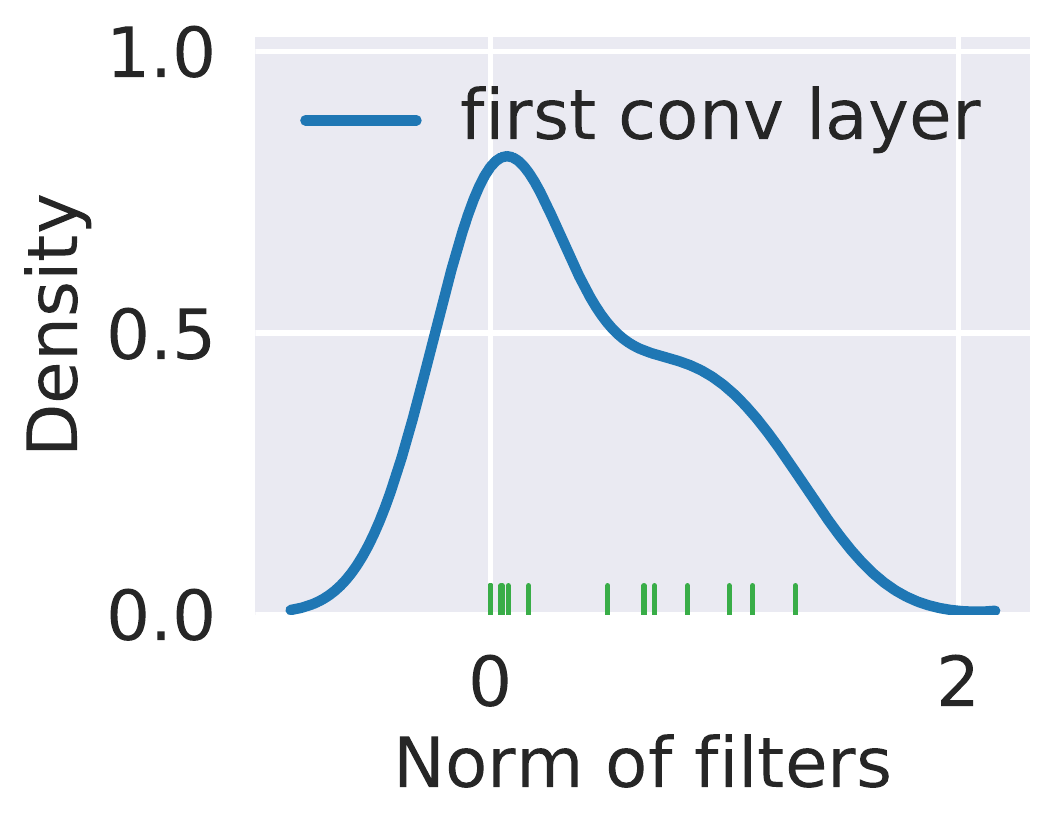}
}
\subfigure[ResNet-110 (log x-scale)]{
\label{fig:norm_dist2}
\includegraphics[width=0.235\linewidth]{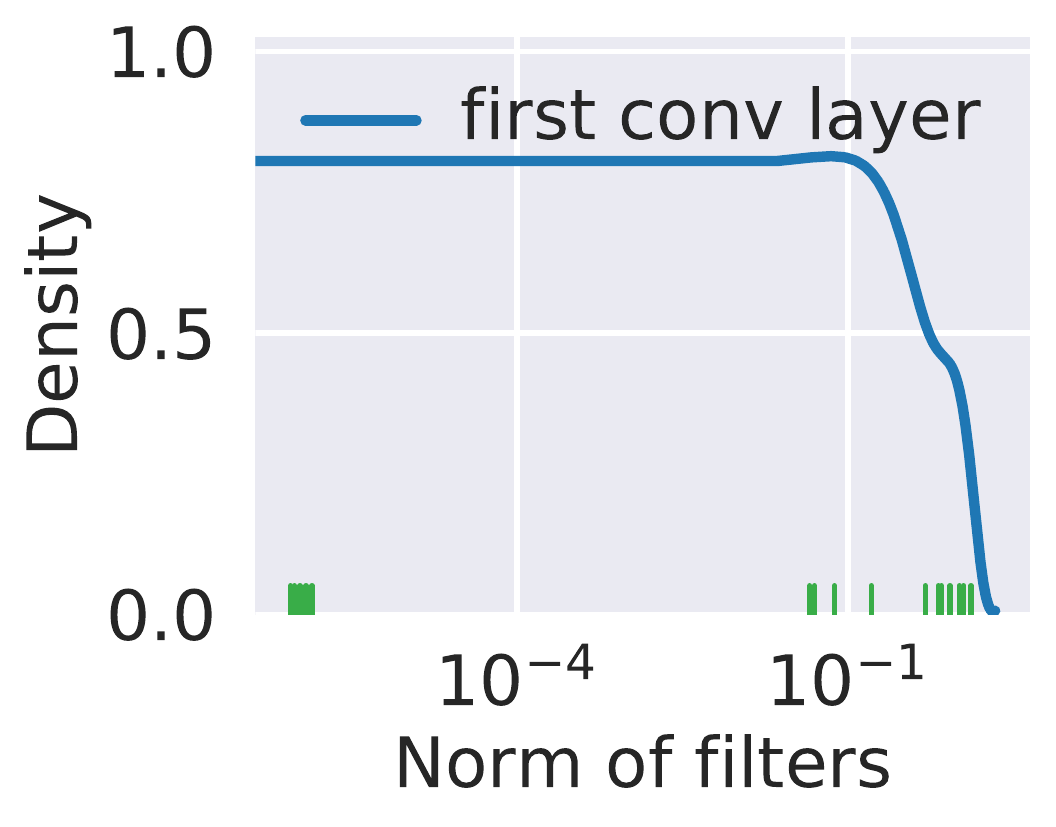}
}
\subfigure[ResNet-110 (linear x-scale)]{
\label{fig:norm_dist3}
\includegraphics[width=0.22\linewidth]{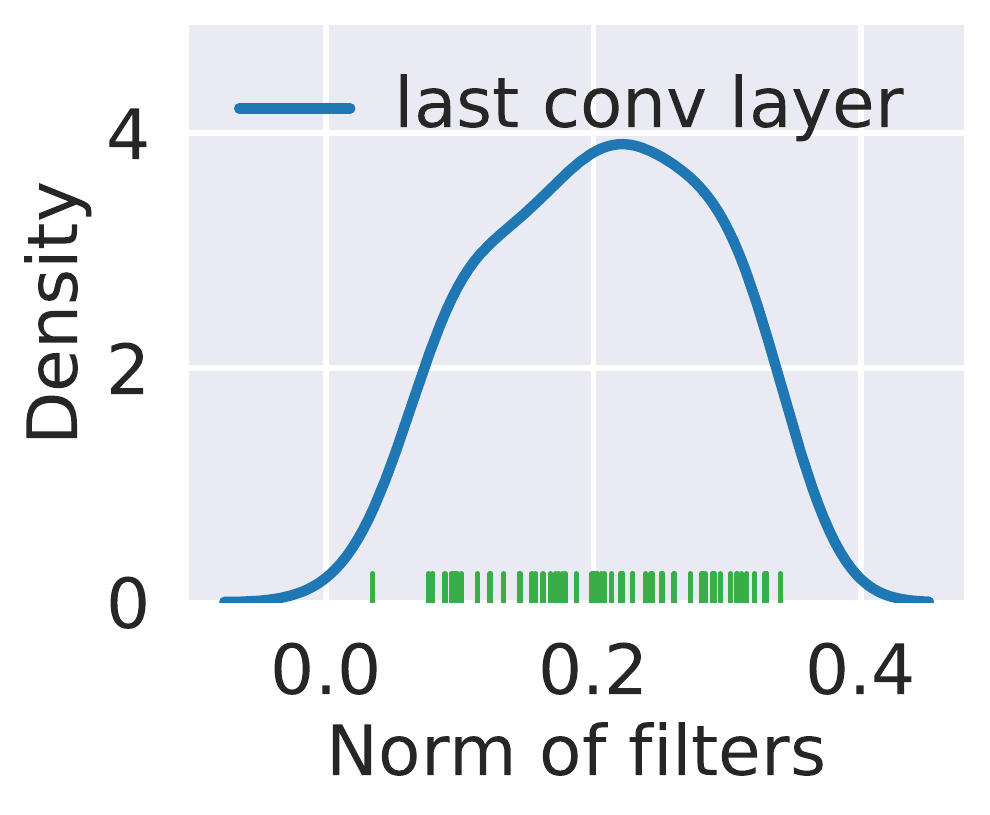}
}
\subfigure[ResNet-110 (log x-scale)]{
\label{fig:norm_dist4}
\includegraphics[width=0.22\linewidth]{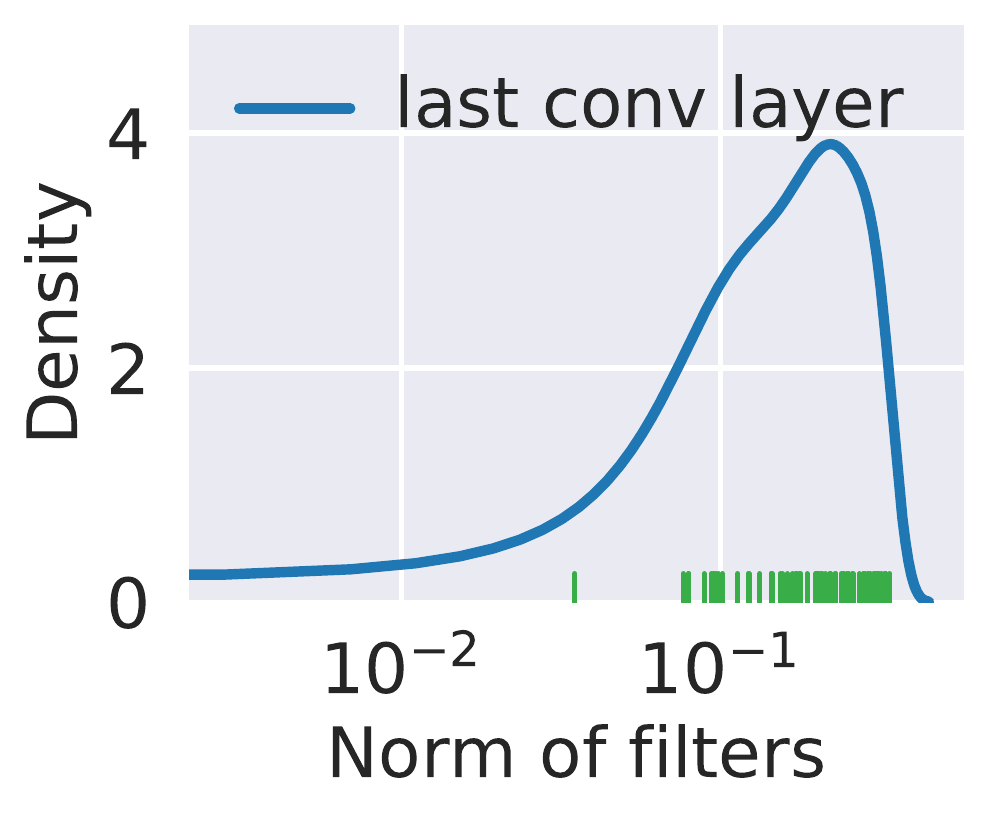}
}
\bigskip
\subfigure[ResNet-18 (linear x-scale)]{
\label{fig:norm_dist5}
\includegraphics[width=0.235\linewidth]{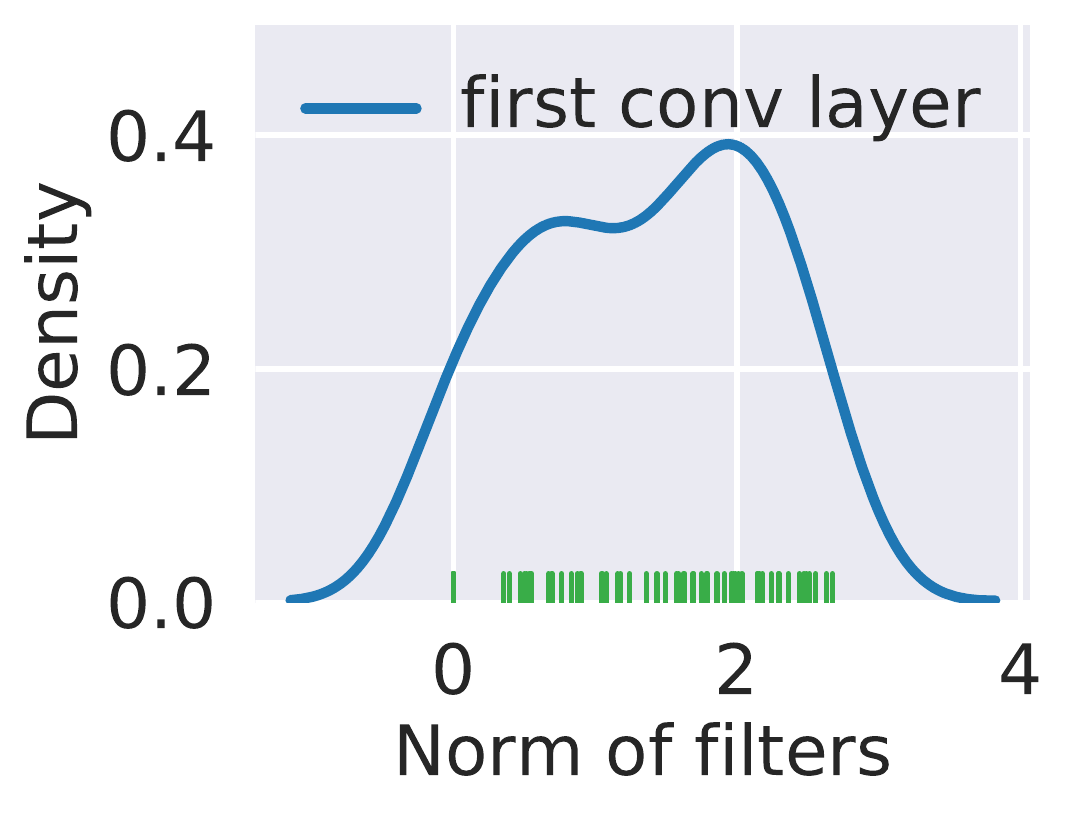}
}
\subfigure[ResNet-18 (log x-scale)]{
\label{fig:norm_dist6}
\includegraphics[width=0.235\linewidth]{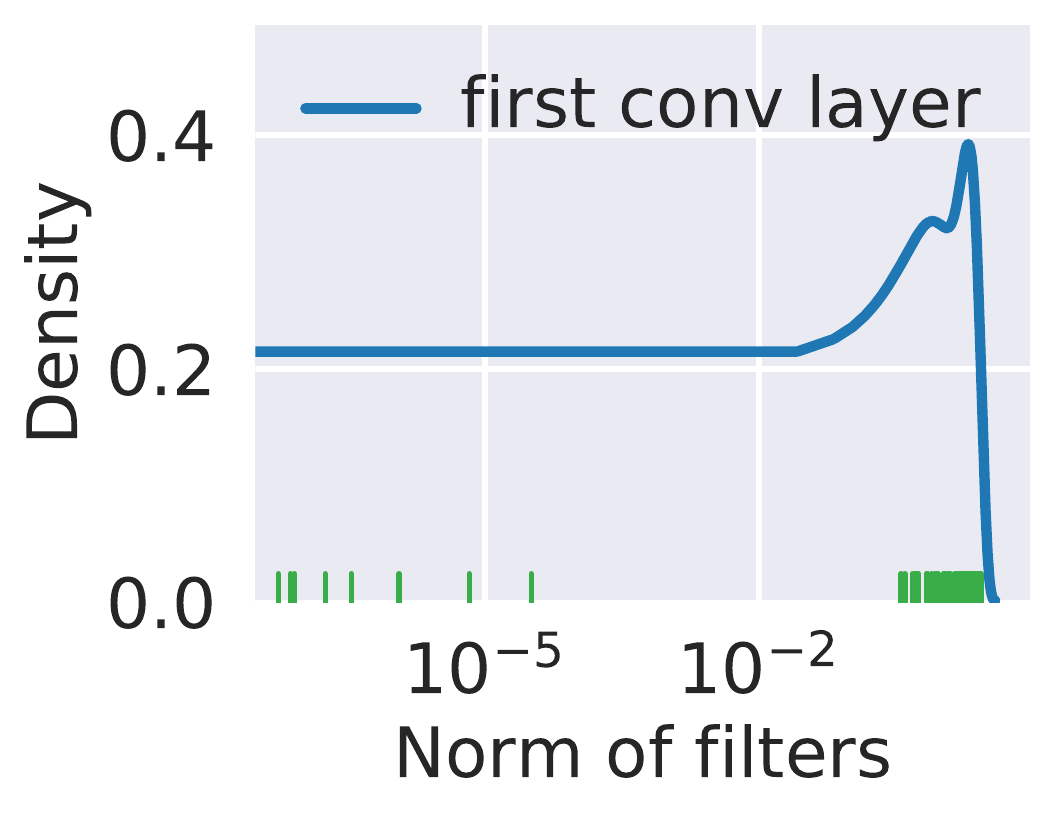}
}
\subfigure[ResNet-18 (linear x-scale)]{
\label{fig:norm_dist7}
\includegraphics[width=0.23\linewidth]{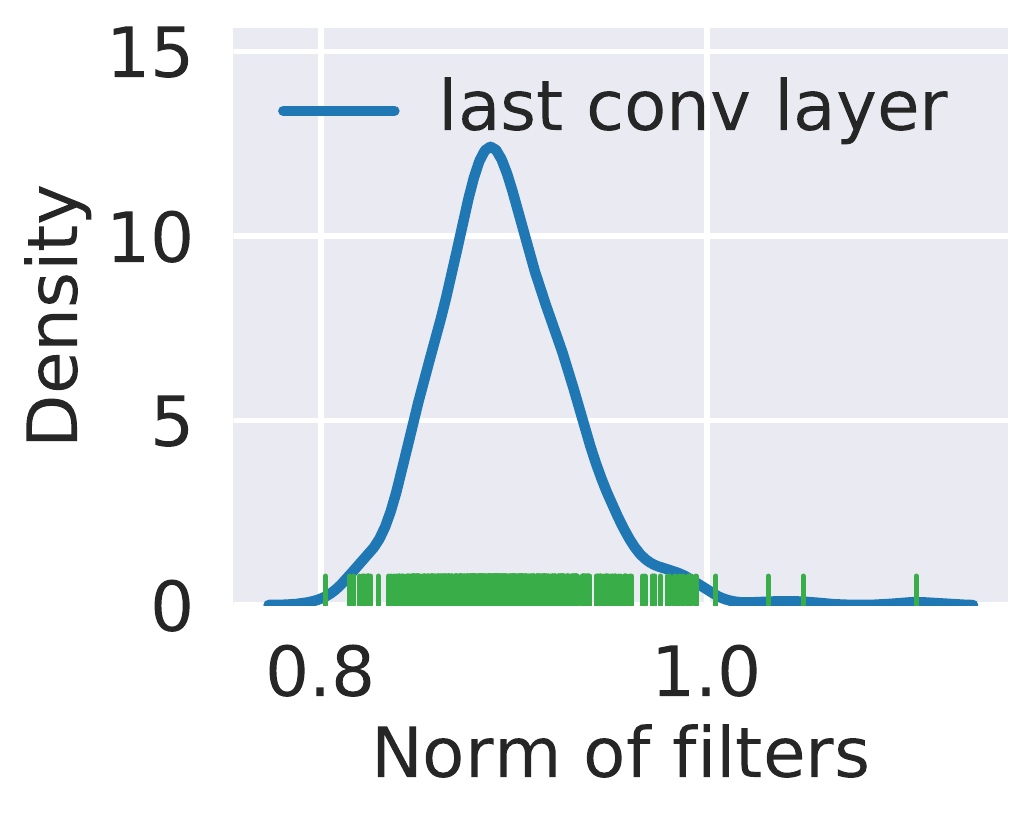}
}
\subfigure[ResNet-18 (log x-scale)]{
\label{fig:norm_dist8}
\includegraphics[width=0.23\linewidth]{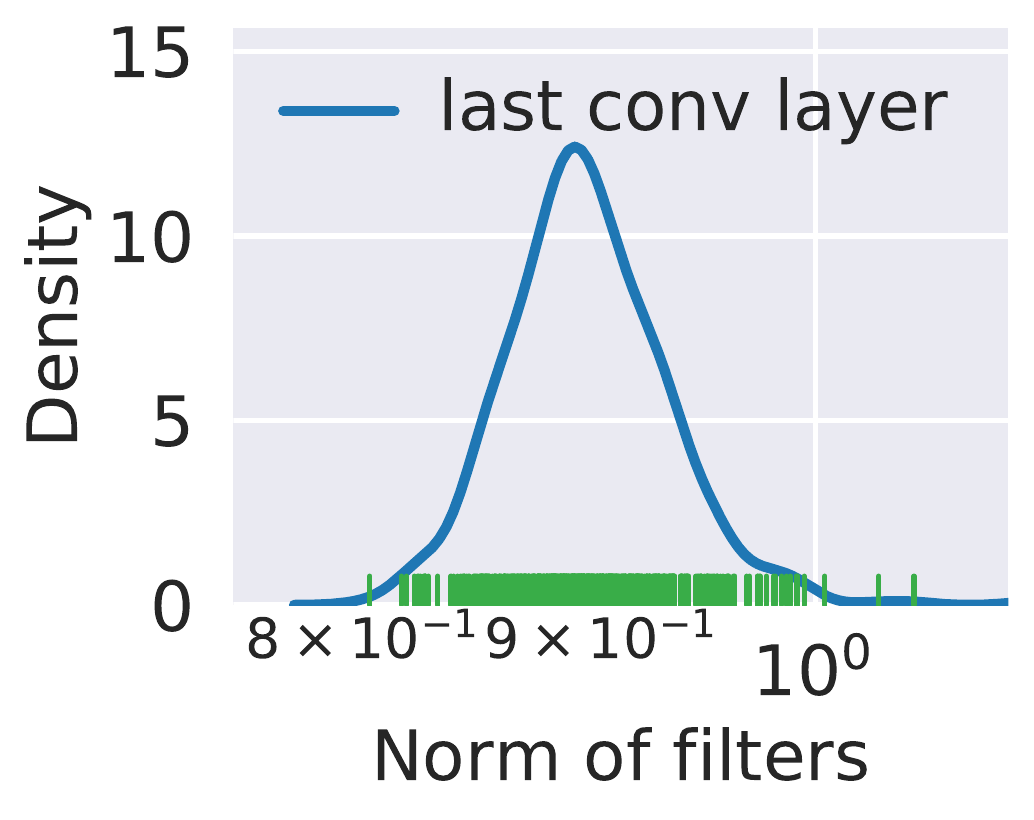}
}
\caption{
Norm distribution of filters from different layers of ResNet-110 on CIFAR-10 and ResNet-18 on ILSVRC-2012. The small green vertical lines and blue curves denote each norm and Kernel Distribution Estimate (KDE) of the norm distribution, respectively.
}
\label{fig:norm_dist}
\end{figure*}

In Figure~\ref{fig:norm_dist}, statistical information collected from pre-trained ResNet-110 on CIFAR-10 and pre-trained ResNet-18 on ILSVRC-2012 demonstrates previous analysis.
The small green vertical lines show each observation in this norm distribution, and the blue curves denote the Kernel Distribution Estimate (KDE)~\cite{silverman2018density}, which is a non-parametric way to estimate the probability density function of a random variable. The norm distribution of first layer and last layer in both structures are drawn. In addition, to clearly illustrate the relation between norm points, two different x-scale, \textit{i.e.}, linear x-scale and log x-scale, are presented.

(1) \emph{Small Norm Deviation in Network}. For the first convolutional layer of ResNet-110, as shown in Figure~\ref{fig:norm_dist}(b), there is a large quantity of filters whose norms are \textbf{concentrated} around the magnitude of \textbf{$10^{-6}$}. 
For the last convolutional layer of ResNet-110, as shown in Figure~\ref{fig:norm_dist}(c), the interval span of the value of norm is roughly \textbf{0.3}, which is much smaller than the interval span of the norm of the first layer (\textbf{1.7}). For the last convolutional layer of ResNet-18, as shown in Figure~\ref{fig:norm_dist}(g), most filter norms are between the interval $[0.8,1.0]$. In all these cases, filters are distributed too densely, which makes it difficult to select a proper threshold to distinguish the important filters from the others. 


(2) \emph{Large Minimum Norm in Network}. For the last convolutional layer of ResNet-18, as shown in Figure~\ref{fig:norm_dist}(g), the minimum norm of these filters is around \textbf{0.8}, \BB{which is \textbf{large} comparing to filters} in the first convolutional layer (Figure~\ref{fig:norm_dist}(e)). For the last convolutional layer of ResNet-110, as shown in Figure~\ref{fig:norm_dist}(c), only one filter is arbitrarily small, while the others are not. Under those circumstances, the filters with minimum norms, although they are relatively less important according to the norm-based criterion, still make significant contributions in the network.

\subsection{Filter Pruning via Geometric Median}
\label{geometric median} 
\BB{To get rid of the constraints in the norm-based criterion, we propose a new filter pruning method inspired from geometric median.}
The central idea of geometric median~\cite{fletcher2008robust} is as follows: given a set of $n$ points $a^{(1)},\ldots,a^{(n)}$ with each $a^{(i)}\in\mathbb{R}^{d}$, find a
point $x^{*}\in\mathbb{R}^{d}$ that minimizes the sum of Euclidean distances to them: 

{\small
\begin{align}
\label{eq:2}
\setlength{\abovedisplayskip}{6pt}
\setlength{\belowdisplayskip}{6pt}
x^{*}=\argmin_{x\in\mathbb{R}^{d}}f(x)\enspace\text{ where }\enspace f(x)\defeq\sum_{i\in[1,n]}\|x-a^{(i)}\|_{2}
\end{align}
}
\RR{where $[1,n]=\{ 1, ..., n \}$.}

\BB{As} the geometric median is a classic robust estimator of centrality for data in Euclidean spaces~\cite{fletcher2008robust}, we use the geometric median to get the common information of all the filters within the single $i_{th}$ layer:
{\small
\begin{equation}
\begin{split}
\label{eq:3}
x^{GM}
&=\argmin_{x\in \mathbb{R}^{N_{i}\times K\times K}} \sum_{j^{'}\in[1,N_{i+1}]}\| x -\mathcal{F}_{i,j^{'}}\|_{2},
\end{split}
\end{equation}
}


In the $i_{th}$ layer, find the filter(s) nearest to the geometric median in that layer:
{\small
\begin{align}
\label{eq:5}
\mathcal{F}_{i,j^*}=\argmin_{\mathcal{F}_{i,j{'}} }\| \mathcal{F}_{i,j^{'}}-x^{GM}
\|_{2}, ~\textrm{s.t.}~j{'}\in[1,N_{i+1}],
\end{align}
}

\noindent 
then $\mathcal{F}_{i,j^*}$ can be represented by the other filters in the same layer, and therefore, pruning them has little negative \BB{impacts} on the network performance.

As geometric median is a non-trivial problem
in computational geometry, the previous fastest running times for computing a $(1 +\epsilon)$-approximate geometric median were $\otilde(dn^{4/3}\cdot\epsilon^{-8/3})$ by~\cite{chin2013runtime}, $O(nd\log^{3}(n/\epsilon))$ by~\cite{cohen2016geometric}, and this is time-consuming. 

In our case, as the final result is in a list of known points, that is, the candidate filters in $i_{th}$ layer. We could instead find which filter minimize\RR{s} the summation of the distance with other filters:


\RR{ 
{\small
\begin{equation}
\begin{split}
\label{eq:7}
\mathcal{F}_{i,x^*}
& \!=\! \argmin_{x }\!\sum_{j^{'}\!\in\![1,N_{i+1}]}\!\!\!\!\|x-\mathcal{F}_{i,j^{'}}\|_{2},~\textrm{s.t.}~x\!\in\!\{ \mathcal{F}_{i,1}, ..., \mathcal{F}_{i,N_{i+1}} \} \\
& \defeq \argmin_{x} g(x), ~\textrm{s.t.}~ x\!\in\!\{ \mathcal{F}_{i,1}, ..., \mathcal{F}_{i,N_{i+1}} \}
\end{split}
\end{equation}
}
}
\noindent

Note that even if $\mathcal{F}_{i,x^*}$ is not included in the calculation of the geometric median in Equation.\ref{eq:7}\footnote{To select multiple filters, we choose several $x$ that makes $g(x)$ to the smallest extent.}, we could also achieve the same result.

In this setting, we want to find the filter

\RR{
{\small
\begin{align}
\label{eq:8}
\mathcal{F}_{i,x^*{'}}=\argmin_{x}g{'}(x),~\textrm{s.t.}~ x\!\in\!\{ \mathcal{F}_{i,1}, ..., \mathcal{F}_{i,N_{i+1}} \}
\end{align}
}
\noindent
where
{\small
\begin{align}
\label{eq:9}
g{'}(x)=\sum_{j{'}\in[1,N_{i+1}], \mathcal{F}_{i,j^{'}} \neq {x} }\|x-\mathcal{F}_{i,j^{'}}\|_{2}.
\end{align}
}
}

\noindent
For each \RR{$x\!\in\!\{ \mathcal{F}_{i,1}, ..., \mathcal{F}_{i,N_{i+1}} \}$}:
\RR{
{\small
\begin{equation}
\begin{split}
g(x) &= 
\sum_{j{'} \in [1,N_{i+1}]} \| x - \mathcal{F}_{i,j{'}} \|_2 \\
& =
\sum_{j{'}\in[1,N_{i+1}], \mathcal{F}_{i,j{'}} \neq x} \| x - \mathcal{F}_{i,j{'}} \|_2
+
[\| x - \mathcal{F}_{i,j{'}} \|_2]_{ \mathcal{F}_{i,j{'}} = x} \\
&=g{'}(x)
\end{split}
\end{equation}
}
}

\noindent
So we could get:
{\small
\begin{equation}
\begin{split}
\label{eq:10}
g(x) &= g{'}(x),~\forall~x \!\in\!\{ \mathcal{F}_{i,1}, ..., \mathcal{F}_{i,N_{i+1}} \}
\end{split}
\end{equation}
}
Thus, we have
\RR{
{\small
\begin{equation}
\begin{split}
\label{eq:11}
\mathcal{F}_{i, x^*} = \argmin_{x \!\in\!\{ \mathcal{F}_{i,1}, ..., \mathcal{F}_{i,N_{i+1}} \} } g(x) &= \argmin_{x \!\in\!\{ \mathcal{F}_{i,1}, ..., \mathcal{F}_{i,N_{i+1}} \} } g{'}(x) = \mathcal{F}_{i, x^*{'}}.
\end{split}
\end{equation}
}
}


Since the geometric median is a classic robust estimator of centrality for data in Euclidean spaces~\cite{fletcher2008robust}, the selected filter(s), $\mathcal{F}_{i,x^{*}}$, and \BB{left ones} share the most common information. This indicates the information of the filter(s) $\mathcal{F}_{i,x^{*}}$ could be replaced by others. After fine-tuning, the network could easily recover its original performance since the information of pruned filters can be represented by the remaining ones. Therefore, the filter(s) $\mathcal{F}_{i,x^{*}}$ could be pruned with negligible effect on the final result of the neural network. The FPGM is summarized in Algorithm~\ref{alg:FPGM}.

\begin{algorithm}[t]
\caption{Algorithm Description of FPGM}
\label{alg:FPGM}

\begin{algorithmic}[1]
\INPUT training data: $\mathbf{X}$.      
\State  \textbf{Given}:  pruning rate $P_{i}$
\State  \textbf{Initialize}: model parameter $\mathbf{W} = \{\mathbf{W} ^{(i)}, 0\leq i \leq L\}$   
\For{$epoch=1$; $epoch \leq epoch_{max}$; $epoch++$}
	\State Update the model parameter $\mathbf{W}$ based on $\mathbf{X}$
	\For{$i=1$; $i \leq L $; $i++$}          
		\State Find  $N_{i+1}P_i$ filters that satisfy Equation~\ref{eq:7}
		\State Zeroize selected filters
	\EndFor
\EndFor
\State Obtain the compact model $\mathbf{W} ^{*}$ from $\mathbf{W}$
\OUTPUT The compact model and its parameters $\mathbf{W} ^{*}$
\end{algorithmic} 
\end{algorithm}



\subsection{Theoretical and Realistic Acceleration}\label{Calculation Reduction After Pruning}

\subsubsection{Theoretical Acceleration}
Suppose the shapes of input tensor $\mathbf{I} \in N_{i} \times H_{i}\times W_{i}$ and output tensor $\mathbf{O} \in N_{i+1} \times H_{i+1}\times W_{i+1}$.
Set the filter pruning rate of the $i_{th}$ layer to $P_{i}$, then $N_{i+1}\times P_{i}$ filters should be pruned.
After filter pruning, the dimension of input and output feature map of the $i_{th}$ layer change to $\mathbf{I'} \in [N_{i}\times(1-P_{i})]\times H_{i}\times W_{i}$ and $\mathbf{O'} \in [N_{i+1}\times(1-P_{i})]\times H_{i+1}\times W_{i+1}$, respectively.

\BB{If setting pruning rate for the $(i+1)_{th}$ layer to $P_{i+1}$, then only $(1-P_{i+1})\times(1-P_{i})$ of the original computation is needed. Finally, a compact model $\{\mathbf{W^{*}}^{(i)}\in\mathbb{R}^{N_{i+1}(1-P_{i})\times N_{i}(1-P_{i-1})\times K\times K}\}$ is obtained.}

\subsubsection{Realistic Acceleration}


\BB{In the above analysis, only the FLOPs of convolution operations for computation complexity comparison is considered, which is common in previous works~\cite{li2016pruning,he2018soft}. This is because other operations such as batch normalization (BN) and pooling are insignificant comparing to convolution operations.}

However, non-tensor layers (e.g., BN and pooling layers) also need the inference time on GPU~\cite{Luo_2017_ICCV}, and influence the realistic acceleration. Besides, the wide gap between the theoretical and realistic acceleration could also be restricted by the IO delay, buffer switch, and efficiency of BLAS libraries. We compare the theoretical and practical acceleration in Table~\ref{table:Comparison_Speed}.

\section{Experiments}\label{Experiment}


\begin{table*}[th]\small
\centering  	
\begin{tabular}{|c|c c c c c c c|}  		
\hline 		
Depth & Method &Fine-tune? &Baseline acc. (\%)  &Accelerated acc. (\%)  &Acc. $\downarrow$ (\%) & FLOPs & FLOPs $\downarrow$(\%)       
\\ \hline \hline   
\multirow{4}{*}{20}        
& SFP~\cite{he2018soft} & \xmark       & \textbf{92.20} ($\pm$0.18)       &  90.83  ($\pm$0.31)& 1.37&  {2.43E7} &{42.2}  \\
& Ours (FPGM-only 30\%) & \xmark       & \textbf{92.20} ($\pm$0.18)       &  {91.09} ($\pm$0.10) & {1.11}&  {2.43E7} &{42.2}  \\  
& Ours (FPGM-only 40\%) & \xmark       & \textbf{92.20} ($\pm$0.18)       &  {90.44} ($\pm$0.20) & {1.76}&  \textbf{1.87E7} &\textbf{54.0}  \\  
& Ours (FPGM-mix 40\%) & \xmark       & \textbf{92.20} ($\pm$0.18)       &  \textbf{90.62} ($\pm$0.17) & \textbf{1.58}&  \textbf{1.87E7} &\textbf{54.0}  \\  

\hline     \hline         

\multirow{5}{*}{32} &MIL~\cite{Dong_2017_CVPR} &\xmark       &92.33         & 90.74 &1.59 &	4.70E7	  & 31.2	 \\          
&SFP~\cite{he2018soft}  & \xmark    & \textbf{92.63} ($\pm$0.70)       &  92.08 ($\pm$0.08)& 0.55 &  {4.03E7} &{41.5} \\ 
& Ours (FPGM-only 30\%)   & \xmark    & \textbf{92.63} ($\pm$0.70)       &  {92.31} ($\pm$0.30) & {0.32} &  {4.03E7} &{41.5} \\   
& Ours (FPGM-only 40\%)   & \xmark    & \textbf{92.63} ($\pm$0.70)       & \textbf{91.93} ($\pm$0.03)& \textbf{0.70} &  \textbf{3.23E7} &\textbf{53.2} \\
& Ours (FPGM-mix 40\%)   & \xmark    & \textbf{92.63} ($\pm$0.70)       &  {91.91} ($\pm$0.21) & {0.72}&  \textbf{3.23E7} &\textbf{53.2} \\

\hline  \hline

\multirow{9}{*}{56} 		
&PFEC~\cite{li2016pruning} &\xmark      & 93.04                   & 91.31  &1.75 &9.09E7  &27.6		 \\         
&CP~\cite{He_2017_ICCV} &\xmark       & 92.80                      & 90.90 &1.90 &	\textbf{--}	  & 50.0	 \\         

&  SFP~\cite{he2018soft}  &\xmark  	&\textbf{93.59} ($\pm$0.58)  & 92.26  ($\pm$0.31)& 1.33   &  \textbf{5.94E7} &\textbf{52.6} \\       
& Ours (FPGM-only 40\%)    &\xmark  	&\textbf{93.59} ($\pm$0.58)  & \textbf{92.93} ($\pm$0.49) & \textbf{0.66}   &  \textbf{5.94E7} &\textbf{52.6} \\   
& Ours (FPGM-mix 40\%)    &\xmark  	&\textbf{93.59} ($\pm$0.58)  & {92.89} ($\pm$0.32) & {0.70}   &  \textbf{5.94E7} &\textbf{52.6} \\   

\cdashline{2-8} 

&PFEC~\cite{li2016pruning} &\cmark      & 93.04          & 93.06 &-0.02 &9.09E7 & 27.6 \\	

&CP~\cite{He_2017_ICCV} &\cmark        & 92.80                   & 91.80 &1.00 &	\textbf{--}	 &	50.0 	 \\

& Ours (FPGM-only 40\%)   &\cmark  	&\textbf{93.59} ($\pm$0.58)   &\textbf{93.49} ($\pm$0.13) &\textbf{0.10}   & \textbf{5.94E7} &\textbf{52.6} \\      
& Ours (FPGM-mix 40\%)   &\cmark  	&\textbf{93.59} ($\pm$0.58)   &{93.26} ($\pm$0.03) &{0.33}   & \textbf{5.94E7} &\textbf{52.6} \\ 

\hline    \hline   

\multirow{8}{*}{110}   
&MIL~\cite{Dong_2017_CVPR} &\xmark        & 93.63         & {93.44} &{0.19} &	-	  & 34.2 	 \\          
&PFEC~\cite{li2016pruning} &\xmark   &93.53     & 92.94      &  0.61              & 1.55E8 	&38.6 	 \\          
& SFP~\cite{he2018soft}    &\xmark  & \textbf{93.68} ($\pm$0.32) 	& 93.38  ($\pm$0.30)& 0.30& {1.50E8} &{40.8}   \\  		
& Ours (FPGM-only 40\%)   &\xmark  & \textbf{93.68} ($\pm$0.32) 	& 93.73  ($\pm$0.23)& -0.05& \textbf{1.21E8} &\textbf{52.3}\\ 
& Ours (FPGM-mix 40\%)   &\xmark  & \textbf{93.68} ($\pm$0.32) 	& \textbf{93.85} ($\pm$0.11) & \textbf{-0.17} & \textbf{1.21E8} &\textbf{52.3}\\ 

\cdashline{2-8} 

&PFEC~\cite{li2016pruning} &\cmark     & 93.53    & 93.30  &{0.20} 	&1.55E8 	&38.6\\ 
&NISP~\cite{yu2018nisp} &\cmark        & \textbf{--}                   & \textbf{--} &0.18 &	\textbf{--}	 &	43.8 	 \\         

& Ours (FPGM-only 40\%)  &\cmark  & \textbf{93.68} ($\pm$0.32) 	& \textbf{93.74} ($\pm$0.10) &\textbf{-0.16}& \textbf{1.21E8} &\textbf{52.3}   \\  		

\hline  	
\end{tabular}  	
\caption{Comparison of pruned ResNet on CIFAR-10.
In ``Fine-tune?'' column, ``\cmark'' and ``\xmark'' indicates whether to use the pre-trained model as initialization or not, respectively.
The ``Acc.~$\downarrow$'' is the accuracy drop between pruned model and the baseline model, the smaller, the better.
} 	
\label{table:cifar10_accuracy} 
\end{table*}  



\setlength{\tabcolsep}{0.30em} 
\begin{table*}[ht] \small 
\centering 	
\begin{tabular}{|c |c c c c c c c c c|} 		
\hline    		
\multirow{3}{*}{Depth}	   & \multirow{3}{*}{Method}  &\multirow{3}{*}{\shortstack {Fine-\\tune?}}  &\multirow{3}{*}{\shortstack {Baseline\\top-1\\acc.(\%)} }  &\multirow{2}{*}{\shortstack {Accelerated\\top-1\\acc.(\%)} } &\multirow{2}{*}{\shortstack{Baseline\\top-5\\acc.(\%)} }   &\multirow{2}{*}{\shortstack {Accelerated\\top-5\\acc.(\%)} }   &\multirow{3}{*}{\shortstack {Top-1 \\acc. $\downarrow$(\%)} }  &\multirow{3}{*}{\shortstack {Top-5\\acc. $\downarrow$(\%)} } & \multirow{3}{*}{\shortstack {FLOPs$\downarrow$(\%)}}    \\
& & & & & & & & & \\
& & & & & & & & & \\
 \hline     \hline     
            
\multirow{6}{*}{18} &MIL~\cite{Dong_2017_CVPR} &\xmark   &69.98 &66.33 & 89.24     & 86.94	&3.65	 &2.30   & 34.6		 \\	       		  
&SFP~\cite{he2018soft} &\xmark &\textbf{70.28}	&{67.10} & \textbf{89.63}  & {87.78}   &{3.18}   & {1.85} & \textbf{41.8} \\

& Ours (FPGM-only 30\%) &\xmark &\textbf{70.28}	&{67.78} & \textbf{89.63}  & {88.01}   &{2.50}   & {1.62} & \textbf{41.8} \\

& Ours (FPGM-mix 30\%) &\xmark &\textbf{70.28}	&\textbf{67.81} & \textbf{89.63}  & \textbf{88.11}   &\textbf{2.47}   & \textbf{1.52} & \textbf{41.8} \\ \cdashline{2-10}
&Ours (FPGM-only 30\%) &\cmark &\textbf{70.28}	&{68.34} & \textbf{89.63}  & \textbf{88.53}   &{1.94}   & \textbf{1.10} & \textbf{41.8} \\

&Ours (FPGM-mix 30\%) &\cmark &\textbf{70.28}	&\textbf{68.41} & \textbf{89.63}  & {88.48}   &\textbf{1.87}   & {1.15} & \textbf{41.8} \\

 \hline     \hline

\multirow{6}{*}{34}	          

&SFP~\cite{he2018soft}	&\xmark 	&\textbf{73.92}		&71.83	&\textbf{91.62}   & 90.33   & 2.09  & 1.29 & \textbf{41.1}    \\    
 
 &  Ours (FPGM-only 30\%)	&\xmark	&\textbf{73.92}		&71.79	&\textbf{91.62}   &\textbf{90.70}   & {2.13}  &\textbf{0.92}   & \textbf{41.1}    \\     
   &  Ours (FPGM-mix 30\%)	&\xmark	&\textbf{73.92}		& \textbf{72.11}	&\textbf{91.62}   & {90.69}  &\textbf{1.81}    &{0.93}    & \textbf{41.1}    \\ 

   \cdashline{2-10} 

 &PFEC~\cite{li2016pruning} &\cmark    & {73.23}          & 72.17 	&\textbf{--}	 &\textbf{--} &\textbf{1.06} 	&\textbf{--} 	& 24.2	 \\ 

 &  Ours (FPGM-only 30\%)	&\cmark &\textbf{73.92}		& {72.54}	&\textbf{91.62}   & \textbf{91.13}  & {1.38}   &\textbf{0.49}   & \textbf{41.1}    \\    
 
 & Ours (FPGM-mix 30\%)	&\cmark 	&\textbf{73.92}		&\textbf{72.63}	&\textbf{91.62}   & {91.08}  &{1.29}    & {0.54}  & \textbf{41.1}    \\    
 \hline      \hline

\multirow{11}{*}{50}	
&SFP~\cite{he2018soft} & \xmark &\textbf{76.15}		&{74.61}		&\textbf{92.87}	  & {92.06} &1.54	 & 0.81 & 41.8  	 \\
& Ours (FPGM-only 30\%) & \xmark  &\textbf{76.15}		&\textbf{75.03}		&\textbf{92.87}	  & \textbf{92.40} &\textbf{1.12} 	 &\textbf{0.47}   &  42.2  	 \\
&  Ours (FPGM-mix 30\%)  & \xmark  &\textbf{76.15}		&{74.94}		&\textbf{92.87}	  & {92.39} &{1.21} 	 & {0.48}  & 42.2  	 \\
&   Ours (FPGM-only 40\%) & \xmark  &\textbf{76.15}		&{74.13}		&\textbf{92.87}	  & {91.94} & {2.02}	 &{0.93}  & \textbf{53.5}  	 \\

\cdashline{2-10} 
&ThiNet~\cite{Luo_2017_ICCV}  &\cmark  &72.88 	&72.04  & 91.14   & 90.67              & {0.84} 	&{0.47}	& 36.7  \\  		  
&SFP~\cite{he2018soft} & \cmark  &\textbf{76.15}		&{62.14}		&\textbf{92.87}	  & {84.60} &14.01	 & 8.27 & 41.8  	 \\
&NISP~\cite{yu2018nisp} & \cmark  &\textbf{--}		&\textbf{--}	&\textbf{--}	  & \textbf{--} & 0.89	 & \textbf{--}& 44.0  	 \\
&CP~\cite{He_2017_ICCV} &\cmark 	&\textbf{--}	&\textbf{--} & 92.20    &90.80  	&\textbf{--}	& 1.40	 &{50.0}	 	\\     		  

&   Ours (FPGM-only 30\%)  & \cmark  &\textbf{76.15}		&\textbf{75.59}		&\textbf{92.87}	  & \textbf{92.63} & \textbf{0.56}	 &{0.24}  & 42.2  	 \\

&  Ours (FPGM-mix 30\%)       &\cmark &\textbf{76.15}		&{75.50}		&\textbf{92.87}	  & {92.63} &{0.65} 	 &\textbf{0.21}  & 42.2  	 \\
&  Ours (FPGM-only 40\%)  & \cmark  &\textbf{76.15}		&{74.83}		&\textbf{92.87}	  & {92.32} & {1.32}	 &{0.55}  &\textbf{53.5}   	 \\

\hline   \hline

\multirow{2}{*}{101}


& Rethinking~\cite{ye2018rethinking}  & \cmark  &\textbf{77.37}		& 75.27 	&\textbf{--}  &  \textbf{--}  & 2.10 	 &  \textbf{--}  & \textbf{47.0}    \\  
& Ours (FPGM-only 30\%) &\cmark  &\textbf{77.37}		&\textbf{77.32}		&\textbf{93.56}	  & \textbf{93.56} &\textbf{0.05}	 & \textbf{0.00}& {42.2}    \\ 		
\hline    	
\end{tabular} 	
\caption{
Comparison of pruned ResNet on ILSVRC-2012.
``Fine-tune?'' and "acc. $\downarrow$" have the same meaning with Table~\ref{table:cifar10_accuracy}.
} 
\label{table:imagenet_accuracy}
\end{table*}


We evaluate FPGM for single-branch network (VGGNet ~\cite{simonyan2014very}), and multiple-branch network (ResNet) on two benchmarks: CIFAR-10~\cite{krizhevsky2009learning} and ILSVRC-2012~\cite{russakovsky2015imagenet}\footnote{As stated in~\cite{li2016pruning}, ``comparing with AlexNet or VGG (on ILSVRC-2012), both VGG (on CIFAR-10) and Residual networks have fewer parameters in the fully connected layers", which makes pruning filters in those networks challenging.}. The CIFAR-10~\cite{krizhevsky2009learning} dataset contains $60,000$ $32 \times 32$ color images in $10$ different classes, in which $50,000$ training images and $10,000$ testing images are included.
ILSVRC-2012~\cite{russakovsky2015imagenet} is a large-scale dataset containing 1.28 million training images and 50k validation images of 1,000 classes.

\subsection{Experimental Settings}
\textbf{Training setting.} On CIFAR-10, the parameter setting is the same as~\cite{he2016identity} and the training schedule is the same as~\cite{zagoruyko2016wide}. 
In the ILSVRC-2012 experiments, we use the default parameter settings which is same as~\cite{he2016deep,he2016identity}.
Data argumentation strategies for ILSVRC-2012 is the same as PyTorch~\cite{paszke2017automatic} official examples.
We analyze the difference between starting from scratch and the pre-trained model. For pruning the model from scratch, We use the normal training schedule without additional fine-tuning process. For pruning the pre-trained model, we reduce the learning rate to one-tenth of the original learning rate. To conduct a fair comparison of pruning scratch and pre-trained models, we use the same training epochs to train/fine-tune the network. The previous work~\cite{li2016pruning} might use fewer epochs to finetune the pruned model, but it converges too early, and its accuracy can not improve even with more epochs, which can be shown in section~\ref{section:vgg}.

\textbf{Pruning setting.} In the filter pruning step, we simply prune \emph{all} the weighted layers with the \emph{same} pruning rate at the same time, which is the same as \cite{he2018soft}.
Therefore, only one hyper-parameter $P_{i}=P$ is needed to balance the acceleration and accuracy. The pruning operation is conducted at the end of every training epoch. Unlike previous work~\cite{li2016pruning}, sensitivity analysis is not essential in FPGM to achieve good performances, which will be demonstrated in later sections.

Apart from FPGM only criterion, we also use a mixture of FPGM and previous norm-based method~\cite{he2018soft} to show that FPGM could serve as a supplement to previous methods. FPGM only criterion is denoted as ``FPGM-only'', the criterion combining the FPGM and norm-based criterion is indicated as ``FPGM-mix''.
``FPGM-only 40\%'' means 40\% filters of the layer are selected with FPGM only, while ``FPGM-mix 40\%'' means 30\% filters of the layer are selected with norm-based criterion~\cite{he2018soft}, and the remaining 10\% filters are selected with FPGM.
We compare FPGM with previous acceleration algorithms, e.g., MIL~\cite{Dong_2017_CVPR}, PFEC~\cite{li2016pruning}, CP~\cite{He_2017_ICCV}, ThiNet~\cite{Luo_2017_ICCV}, SFP~\cite{he2018soft}, NISP~\cite{yu2018nisp}, Rethinking~\cite{ye2018rethinking}. Not surprisingly, our FPGM method achieves the state-of-the-art result.

\subsection{Single-Branch Network Pruning}
\label{section:vgg}
\textbf{VGGNet on CIFAR-10.}
As the training setup is not publicly available for~\cite{li2016pruning}, we re-implement the pruning procedure and achieve similar results to the original paper. The result of pruning pre-trained and scratch model is shown in Table~\ref{table:vgg_pretrain} and Table~\ref{table:vgg_scratch}, respectively. Not surprisingly, FPGM achieves better performance than ~\cite{li2016pruning} in both settings.

\begin{table}[ht]
\small
\setlength{\tabcolsep}{0.25em}
\begin{center}
\begin{tabular}{| c | c | c | c  | c |}
\hline
\multirow{2}{*}{Model $\backslash$ Acc (\%)} & \multirow{2}{*}{\shortstack{Baseline}} & \multirow{2}{*}{\shortstack {Pruned\\w.o. FT}} & \multirow{2}{*}{\shortstack { FT  \\40 epochs}} &\multirow{2}{*}{\shortstack {FT \\160 epochs} }\\
               &         &          &        &          \\ \hline
               
\multirow{2}{*} {PFEC~\cite{li2016pruning}  }    & \multirow{2}{*} { \shortstack{93.58 \\($\pm$0.03)} }  & \multirow{2}{*} { \shortstack{77.45\\($\pm$0.03)} } &\multirow{2}{*} {\shortstack{93.22  \\($\pm$0.03 )}}   &  \multirow{2}{*} {\shortstack{93.28 \\ ($\pm$0.07)}  }  \\  
               &         &          &        &          \\ \hline
               
\multirow{2}{*} {Ours}    & \multirow{2}{*} { \shortstack{93.58 \\($\pm$0.03)} }  & \multirow{2}{*} { \shortstack{\textbf{80.38}\\($\pm$0.03)} } &\multirow{2}{*} {\shortstack{\textbf{93.24}  \\($\pm$0.01) }}   &  \multirow{2}{*} {\shortstack{\textbf{94.00}\\ ($\pm$0.13)}  }  \\  
               &         &          &        &          \\ \hline
\end{tabular}
\end{center}
\caption{Pruning pre-trained VGGNet on CIFAR-10. ``w.o." means ``without" and ``FT" means ``fine-tuning" the pruned model.}

\label{table:vgg_pretrain}
\end{table}

\begin{table}[ht]
\small
\setlength{\tabcolsep}{0.1em}
\begin{center}
\begin{tabular}{|c|c|c|c|c|}
\hline
{Model } & {SA}& {Baseline }         & {Pruned From Scratch } &{FLOPs$\downarrow$(\%)}    \\ \hline
PFEC~\cite{li2016pruning}     &Y  & {93.58 ($\pm$0.03)} & 93.31 ($\pm$0.02)    &{34.2}       \\ \hline
Ours                          &Y  & 93.58 ($\pm$0.03)   & \textbf{93.54} ($\pm$0.08) &{34.2} \\ \hline
Ours                          &N  & 93.58 ($\pm$0.03)   & {93.23} ($\pm$0.13) &\textbf{35.9} \\ \hline

\end{tabular}
\end{center}
\caption{Pruning scratch VGGNet on CIFAR-10. ``SA" means ``sensitivity analysis". Without sensitivity analysis, FPGM can still achieve comparable performances comparing to~\cite{li2016pruning}; after introducing sensitivity analysis, FPGM can surpass~\cite{li2016pruning}.
}
\label{table:vgg_scratch}
\end{table}

\subsection{Multiple-Branch Network Pruning}\label{section:ILSVRC}

\textbf{ResNet on CIFAR-10.} For the CIFAR-10 dataset, we test our FPGM on ResNet-20, 32, 56 and 110 with two different pruning rates: 30\% and 40\%. 

As shown in Table~\ref{table:cifar10_accuracy}, our FPGM achieves the state-of-the-art performance. For example, MIL~\cite{Dong_2017_CVPR} without fine-tuning accelerates ResNet-32 by 31.2\% speedup ratio with 1.59\% accuracy drop, but our FPGM without fine-tuning achieves 53.2\% speedup ratio with even 0.19\% accuracy improvement. Comparing to SFP~\cite{he2018soft}, when pruning 52.6\% FLOPs of ResNet-56, our FPGM has only 0.66\% accuracy drop, which is much less than SFP~\cite{he2018soft} (1.33\%). For pruning the pre-trained ResNet-110, our method achieves a much higher (52.3\% v.s. 38.6\%) acceleration ratio with 0.16\% performance increase, while PFEC~\cite{li2016pruning} harms the performance with lower acceleration ratio.
These results demonstrate that FPGM can produce a more compressed model with comparable or even better performances. 




\begin{table}[ht]
\small
\setlength{\tabcolsep}{0.3em}
\begin{center}
\begin{tabular}{| c | c | c | c  | c |}
\hline
\multirow{2}{*}{Model} & \multirow{2}{*}{\shortstack{Baseline\\time (ms)}} & \multirow{2}{*}{\shortstack {Pruned\\time (ms) }} & \multirow{2}{*}{\shortstack {Realistic \\Acc.(\%)}} &\multirow{2}{*}{\shortstack {Theoretical\\Acc.(\%)} }\\
               &         &          &        &          \\ \hline
ResNet-18      & 37.05   &  26.77   & 27.7   &  41.8    \\  
ResNet-34      & 63.89   &  45.24   & 29.2   &  41.1    \\  
ResNet-50      & 134.57  &  83.22   & 38.2   &  53.5    \\ 
ResNet-101     & 219.70  & 147.45   & 32.9   &  42.2    \\ \hline
\end{tabular}
\end{center}
\caption{
Comparison on the theoretical and realistic acceleration.
Only the time consumption of the forward procedure is considered.
}
\label{table:Comparison_Speed}
\end{table}
\textbf{ResNet on ILSVRC-2012.} For the ILSVRC-2012 dataset, we test our FPGM on ResNet-18, 34, 50 and 101 with pruning rates 30\% and 40\%.
Same with~\cite{he2018soft}, we do not prune the projection shortcuts for simplification. 

Table~\ref{table:imagenet_accuracy} shows that FPGM outperforms previous methods on ILSVRC-2012 dataset, again.
For ResNet-18, pure FPGM without fine-tuning achieves the same inference speedup with~\cite{he2018soft}, but its accuracy exceeds by 0.68\%. FPGM-only with fine-tuning could even gain 0.60\% improvement over FPGM-only without fine-tuning, thus exceeds~\cite{he2018soft} by 1.28\%.
For ResNet-50, FPGM with fine-tuning achieves more inference speedup than CP~\cite{He_2017_ICCV}, but our pruned model exceeds their model by 0.85\% on the accuracy.
Moreover, for pruning a pre-trained ResNet-101, FPGM reduces more than 40\% FLOPs of the model without top-5 accuracy loss and only negligible (0.05\%) top-1 accuracy loss. In contrast, the performance degradation is 2.10\% for Rethinking~\cite{ye2018rethinking}.
Compared to the norm-based criterion, Geometric Median (GM) explicitly utilizes the relationship between filters, which is the main cause of its superior performance.

To compare the theoretical and realistic acceleration, we measure the forward time of the pruned models on one GTX1080 GPU with a batch size of $64$. The results~\footnote{Optimization of the addition of ResNet shortcuts and convolutional outputs would also affect the results.} are shown in Table~\ref{table:Comparison_Speed}.
As discussed in the above section, the gap between the theoretical and realistic model may come from the limitation of IO delay, buffer switch, and efficiency of BLAS libraries.

\subsection{Ablation Study}\label{1-norm and 2-norm}

\begin{figure}[t]
\center
\subfigure[Different pruning intervals]{
\label{fig:different_epoch}
\includegraphics[width=0.48\linewidth]{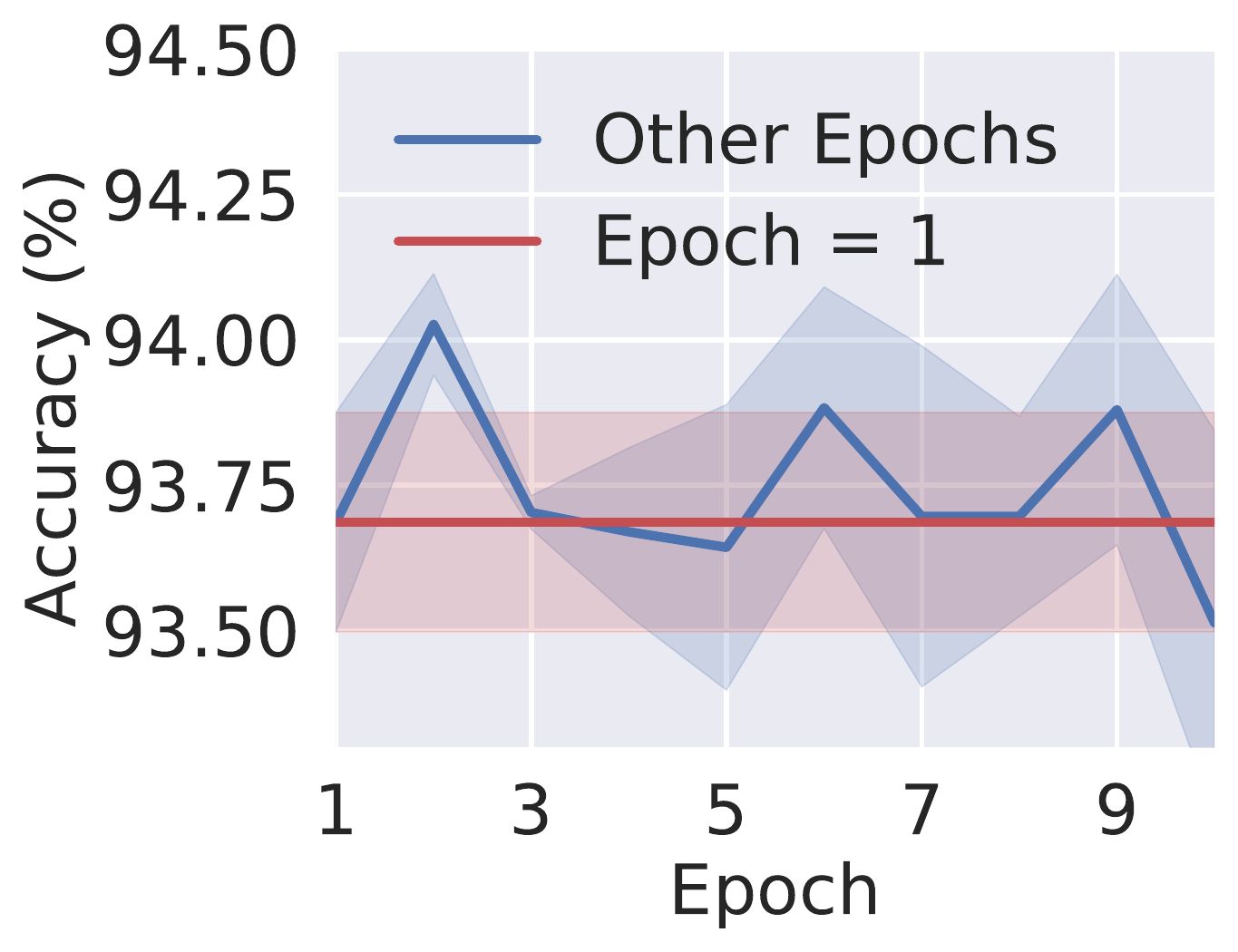}
}
\subfigure[Different pruned FLOPs]{
\label{fig:different_rate}
\includegraphics[width=0.44\linewidth]{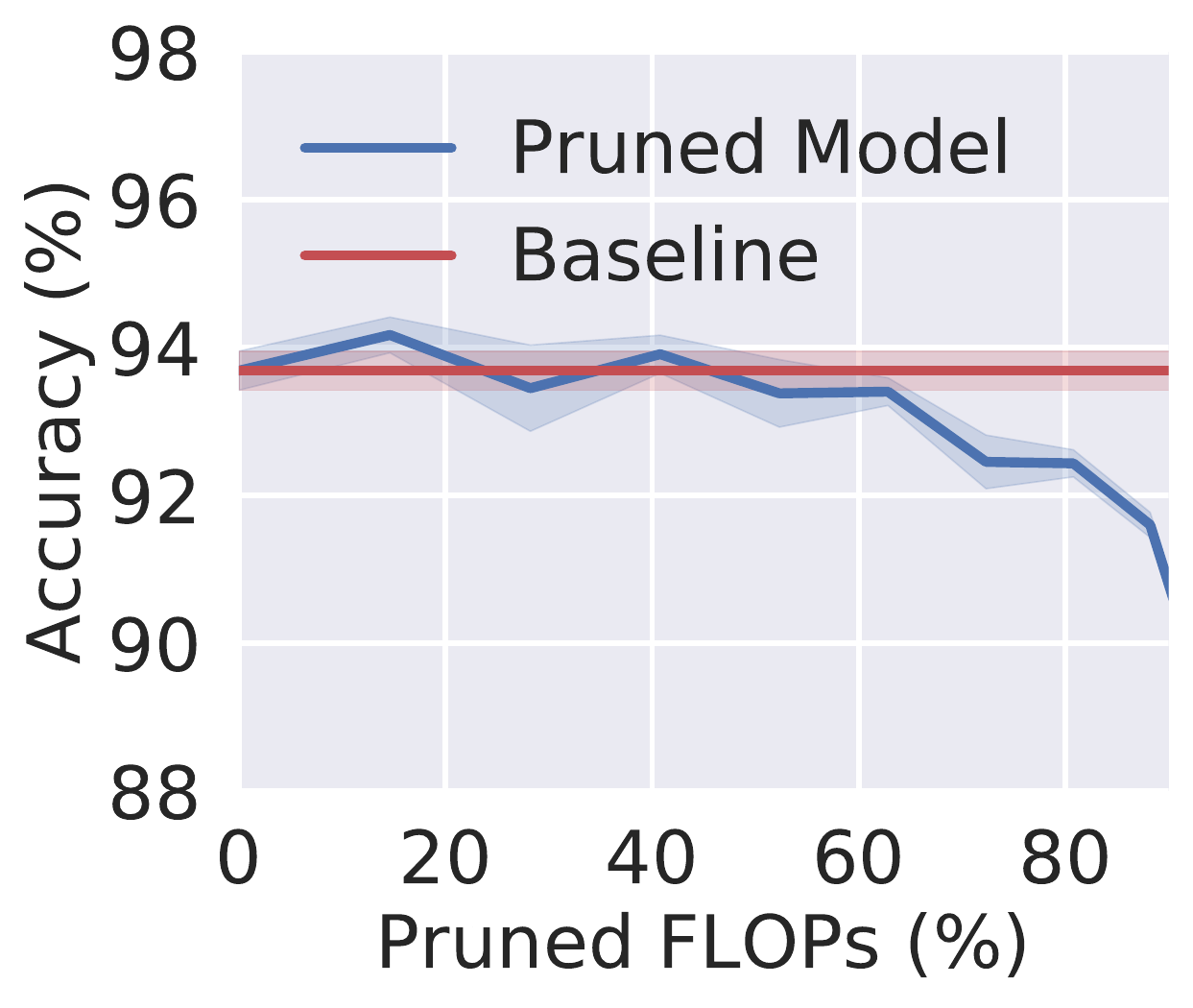}
}
\caption{
Accuracy of ResNet-110 on CIFAR-10 regarding different hyper-parameters. Solid line and shadow denotes the mean values and standard deviation of three experiments, respectively.
}
\label{fig:resnet_cifar10}
\end{figure}

\textbf{Influence of Pruning Interval}
In our experiment setting, the interval of pruning equals to one, \textit{i.e.}, we conduct our pruning operation at the end of every training epoch. To explore the influence of pruning interval, we change the pruning interval from one epoch to ten epochs.
We use the ResNet-110 under pruning rate 40\% as the baseline, as shown in Fig.~\ref{fig:different_epoch}.
The accuracy fluctuation along with the different pruning intervals is less than 0.3\%, which means the performance of pruning is not sensitive to this parameter. Note that fine-tuning this parameter could even achieve better performance.

\textbf{Varying Pruned FLOPs}
We change the ratio of Pruned FLOPs for ResNet-110 to comprehensively understand FPGM, as shown in Fig.~\ref{fig:different_rate}.
When the pruned FLOPs is 18\% and 40\%, the performance of the pruned model even exceeds the baseline model without pruning, which shows FPGM may have a regularization effect on the neural network.

\textbf{Influence of Distance Type}
We use $\ell_{1}$-norm and cosine distance to replace the distance function in Equation~\ref{eq:7}. We use the ResNet-110 under pruning rate 40\% as the baseline, the accuracy of the pruned model is 93.73 $\pm$ 0.23 \%. The accuracy based on $\ell_{1}$-norm and cosine distance is 93.87 $\pm$ 0.22 \% and 93.56 $\pm$ 0.13, respectively. Using $\ell_{1}$-norm as the distance of filter would bring a slightly better result, but cosine distance as distance would slightly harm the performance of the network.

\textbf{Combining FPGM with Norm-based Criterion}
We analyze the effect of combining FPGM and previous norm-based criterion.
For ResNet-110 on CIFAR-10, FPGM-mix is slightly better than FPGM-only. For ResNet-18 on ILSVRC-2012, the performances of FPGM-only and FPGM-mix are almost the same. It seems that the norm-based criterion and FPGM together can boost the performance on CIFAR-10, but not on ILSVRC-2012. We believe that this is because the two requirements for the norm-based criterion are met on some layers of CIFAR-10 pre-trained network, but not on that of ILSVRC-2012 pre-trained network, which is shown in Figure~\ref{fig:norm_dist}.

\subsection{Feature Map Visualization}
We visualize the feature maps of the first layer of the first block of ResNet-50. The feature maps with red titles (7,23,27,46,56,58) correspond to the selected filter activation when setting the pruning rate to 10\%. These selected feature maps contain outlines of the bamboo and the panda's head and body, which can be replaced by remaining feature maps: (5,12,16,18,22,~\etal) containing outlines of the bamboo, and (0,4,33,34,47,~\etal) containing the outline of panda.

\begin{figure}[!ht]
\center
\includegraphics[width=1.0\linewidth]{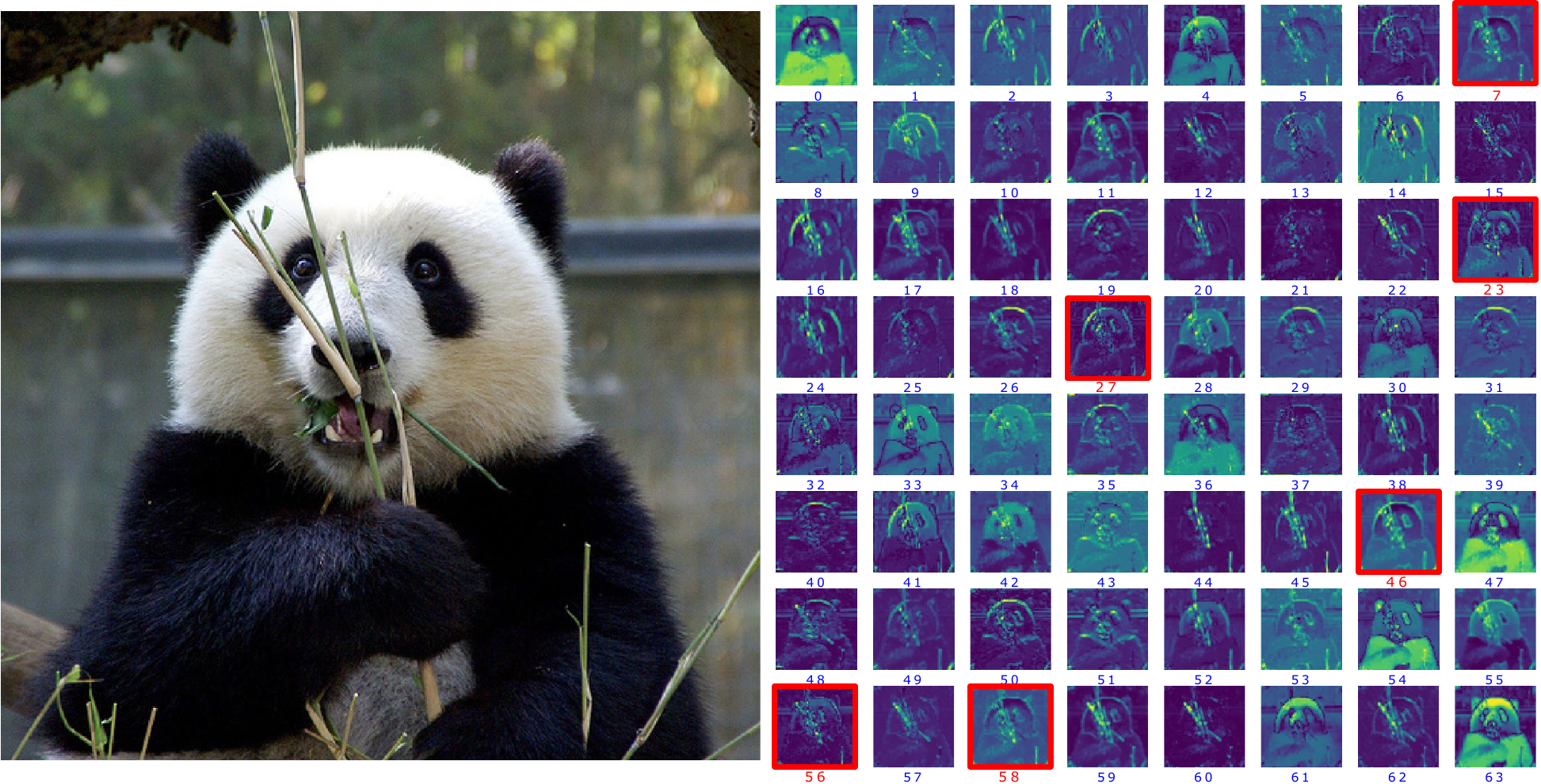}
\caption{
Input image (left) and visualization of feature maps (right) of ResNet-50-conv1. Feature maps with red bounding boxes are the channels to be pruned.
}
\label{fig:visual}
\end{figure}



\section{Conclusion and Future Work}
 
In this paper, we elaborate on the underlying requirements for norm-based filter pruning criterion and point out their limitations. To solve this, we propose a new filter pruning strategy based on the geometric median, named FPGM, to accelerate the deep CNNs. Unlike the previous norm-based criterion, FPGM explicitly considers the mutual relations between filters. Thanks to this, FPGM achieves the state-of-the-art performance in several benchmarks.
In the future, we plan to work on how to combine FPGM with other acceleration algorithms, e.g., matrix decomposition and low-precision weights, to push the performance to a higher stage.

{\small
\bibliographystyle{ieee}
\bibliography{ref}

\begin{thebibliography}{10}\itemsep=-1pt

\bibitem{carreira2018learning}
M.~A. Carreira-Perpin{\'a}n and Y.~Idelbayev.
\newblock “learning-compression” algorithms for neural net pruning.
\newblock In {\em CVPR}, 2018.

\bibitem{chin2013runtime}
H.~H. Chin, A.~Madry, G.~L. Miller, and R.~Peng.
\newblock Runtime guarantees for regression problems.
\newblock In {\em Proceedings of the 4th conference on Innovations in
  Theoretical Computer Science}, pages 269--282. ACM, 2013.

\bibitem{cohen2016geometric}
M.~B. Cohen, Y.~T. Lee, G.~Miller, J.~Pachocki, and A.~Sidford.
\newblock Geometric median in nearly linear time.
\newblock In {\em Proceedings of the forty-eighth annual ACM symposium on
  Theory of Computing}, pages 9--21. ACM, 2016.

\bibitem{dong2017learning}
X.~Dong, S.~Chen, and S.~Pan.
\newblock Learning to prune deep neural networks via layer-wise optimal brain
  surgeon.
\newblock In {\em Advances in Neural Information Processing Systems}, pages
  4857--4867, 2017.

\bibitem{Dong_2017_CVPR}
X.~Dong, J.~Huang, Y.~Yang, and S.~Yan.
\newblock More is less: A more complicated network with less inference
  complexity.
\newblock In {\em CVPR}, 2017.

\bibitem{dong2019search}
X.~Dong and Y.~Yang.
\newblock Searching for a robust neural architecture in four gpu hours.
\newblock In {\em Proceedings of the IEEE Conference on Computer Vision and
  Pattern Recognition (CVPR)}, 2019.

\bibitem{dubey2018coreset}
A.~Dubey, M.~Chatterjee, and N.~Ahuja.
\newblock Coreset-based neural network compression.
\newblock In {\em ECCV}, 2018.

\bibitem{fletcher2008robust}
P.~T. Fletcher, S.~Venkatasubramanian, and S.~Joshi.
\newblock Robust statistics on riemannian manifolds via the geometric median.
\newblock In {\em CVPR}, 2008.

\bibitem{guo2016dynamic}
Y.~Guo, A.~Yao, and Y.~Chen.
\newblock Dynamic network surgery for efficient {DNNs}.
\newblock In {\em NIPS}, 2016.

\bibitem{han2015deep}
S.~Han, H.~Mao, and W.~J. Dally.
\newblock Deep compression: Compressing deep neural networks with pruning,
  trained quantization and huffman coding.
\newblock In {\em ICLR}, 2015.

\bibitem{han2015learning}
S.~Han, J.~Pool, J.~Tran, and W.~Dally.
\newblock Learning both weights and connections for efficient neural network.
\newblock In {\em NIPS}, 2015.

\bibitem{he2016deep}
K.~He, X.~Zhang, S.~Ren, and J.~Sun.
\newblock Deep residual learning for image recognition.
\newblock In {\em CVPR}, 2016.

\bibitem{he2016identity}
K.~He, X.~Zhang, S.~Ren, and J.~Sun.
\newblock Identity mappings in deep residual networks.
\newblock In {\em ECCV}, 2016.

\bibitem{he2018adc}
Y.~He and S.~Han.
\newblock {ADC}: Automated deep compression and acceleration with reinforcement
  learning.
\newblock {\em arXiv preprint arXiv:1802.03494}, 2018.

\bibitem{he2018soft}
Y.~He, G.~Kang, X.~Dong, Y.~Fu, and Y.~Yang.
\newblock Soft filter pruning for accelerating deep convolutional neural
  networks.
\newblock In {\em IJCAI}, 2018.

\bibitem{He_2017_ICCV}
Y.~He, X.~Zhang, and J.~Sun.
\newblock Channel pruning for accelerating very deep neural networks.
\newblock In {\em ICCV}, 2017.

\bibitem{hinton2015distilling}
G.~Hinton, O.~Vinyals, and J.~Dean.
\newblock Distilling the knowledge in a neural network.
\newblock In {\em NIPS}, 2015.

\bibitem{huang2018learning}
Q.~Huang, K.~Zhou, S.~You, and U.~Neumann.
\newblock Learning to prune filters in convolutional neural networks.
\newblock In {\em WACV}, 2018.

\bibitem{kim2018paraphrasing}
J.~Kim, S.~Park, and N.~Kwak.
\newblock Paraphrasing complex network: Network compression via factor
  transfer.
\newblock In {\em NIPS}, 2018.

\bibitem{krizhevsky2009learning}
A.~Krizhevsky and G.~Hinton.
\newblock Learning multiple layers of features from tiny images.
\newblock 2009.

\bibitem{li2016pruning}
H.~Li, A.~Kadav, I.~Durdanovic, H.~Samet, and H.~P. Graf.
\newblock Pruning filters for efficient {ConvNets}.
\newblock In {\em ICLR}, 2017.

\bibitem{lin2019towards}
S.~Lin, R.~Ji, C.~Yan, B.~Zhang, L.~Cao, Q.~Ye, F.~Huang, and D.~Doermann.
\newblock Towards optimal structured cnn pruning via generative adversarial
  learning.
\newblock In {\em CVPR}, 2019.

\bibitem{Liu_2017_ICCV}
Z.~Liu, J.~Li, Z.~Shen, G.~Huang, S.~Yan, and C.~Zhang.
\newblock Learning efficient convolutional networks through network slimming.
\newblock In {\em ICCV}, 2017.

\bibitem{liu2018frequency}
Z.~Liu, J.~Xu, X.~Peng, and R.~Xiong.
\newblock Frequency-domain dynamic pruning for convolutional neural networks.
\newblock In {\em NIPS}, 2018.

\bibitem{Luo_2017_ICCV}
J.-H. Luo, J.~Wu, and W.~Lin.
\newblock {ThiNet}: A filter level pruning method for deep neural network
  compression.
\newblock In {\em ICCV}, 2017.

\bibitem{Yawei2019Taking}
Y.~Luo, L.~Zheng, T.~Guan, J.~Yu, and Y.~Yang.
\newblock Taking a closer look at domain shift: Category-level adversaries for
  semantics consistent domain adaptation.
\newblock In {\em CVPR}, 2019.

\bibitem{molchanov2016pruning}
P.~Molchanov, S.~Tyree, T.~Karras, T.~Aila, and J.~Kautz.
\newblock Pruning convolutional neural networks for resource efficient transfer
  learning.
\newblock In {\em ICLR}, 2017.

\bibitem{paszke2017automatic}
A.~Paszke, S.~Gross, S.~Chintala, G.~Chanan, E.~Yang, Z.~DeVito, Z.~Lin,
  A.~Desmaison, L.~Antiga, and A.~Lerer.
\newblock Automatic differentiation in pytorch.
\newblock In {\em NIPS-W}, 2017.

\bibitem{russakovsky2015imagenet}
O.~Russakovsky, J.~Deng, H.~Su, J.~Krause, S.~Satheesh, S.~Ma, Z.~Huang,
  A.~Karpathy, A.~Khosla, M.~Bernstein, et~al.
\newblock {ImageNet} large scale visual recognition challenge.
\newblock {\em IJCV}, 2015.

\bibitem{silverman2018density}
B.~W. Silverman.
\newblock {\em Density estimation for statistics and data analysis}.
\newblock Routledge, 2018.

\bibitem{simonyan2014very}
K.~Simonyan and A.~Zisserman.
\newblock Very deep convolutional networks for large-scale image recognition.
\newblock In {\em ICLR}, 2015.

\bibitem{Son_2018_ECCV}
S.~Son, S.~Nah, and K.~Mu~Lee.
\newblock Clustering convolutional kernels to compress deep neural networks.
\newblock In {\em The European Conference on Computer Vision (ECCV)}, 2018.

\bibitem{suau2018principal}
X.~Suau, L.~Zappella, V.~Palakkode, and N.~Apostoloff.
\newblock Principal filter analysis for guided network compression.
\newblock {\em arXiv preprint arXiv:1807.10585}, 2018.

\bibitem{szegedy2015going}
C.~Szegedy, W.~Liu, Y.~Jia, P.~Sermanet, S.~Reed, D.~Anguelov, D.~Erhan,
  V.~Vanhoucke, and A.~Rabinovich.
\newblock Going deeper with convolutions.
\newblock In {\em CVPR}, 2015.

\bibitem{tai2015convolutional}
C.~Tai, T.~Xiao, Y.~Zhang, X.~Wang, et~al.
\newblock Convolutional neural networks with low-rank regularization.
\newblock In {\em ICLR}, 2016.

\bibitem{tung2018clip}
F.~Tung and G.~Mori.
\newblock Clip-q: Deep network compression learning by in-parallel
  pruning-quantization.
\newblock In {\em CVPR}, 2018.

\bibitem{wang2018exploring}
D.~Wang, L.~Zhou, X.~Zhang, X.~Bai, and J.~Zhou.
\newblock Exploring linear relationship in feature map subspace for convnets
  compression.
\newblock {\em arXiv preprint arXiv:1803.05729}, 2018.

\bibitem{ye2018rethinking}
J.~Ye, X.~Lu, Z.~Lin, and J.~Z. Wang.
\newblock Rethinking the smaller-norm-less-informative assumption in channel
  pruning of convolution layers.
\newblock In {\em ICLR}, 2018.

\bibitem{yu2018nisp}
R.~Yu, A.~Li, C.-F. Chen, J.-H. Lai, V.~I. Morariu, X.~Han, M.~Gao, C.-Y. Lin,
  and L.~S. Davis.
\newblock {NISP}: Pruning networks using neuron importance score propagation.
\newblock In {\em CVPR}, 2018.

\bibitem{zagoruyko2016wide}
S.~Zagoruyko and N.~Komodakis.
\newblock Wide residual networks.
\newblock In {\em BMVC}, 2016.

\bibitem{zhang2018systematic}
T.~Zhang, S.~Ye, K.~Zhang, J.~Tang, W.~Wen, M.~Fardad, and Y.~Wang.
\newblock A systematic dnn weight pruning framework using alternating direction
  method of multipliers.
\newblock {\em arXiv preprint arXiv:1804.03294}, 2018.

\bibitem{zhang2016accelerating}
X.~Zhang, J.~Zou, K.~He, and J.~Sun.
\newblock Accelerating very deep convolutional networks for classification and
  detection.
\newblock {\em IEEE T-PAMI}, 2016.

\bibitem{zhou2017incremental}
A.~Zhou, A.~Yao, Y.~Guo, L.~Xu, and Y.~Chen.
\newblock Incremental network quantization: Towards lossless cnns with
  low-precision weights.
\newblock In {\em ICLR}, 2017.

\bibitem{zhu2016trained}
C.~Zhu, S.~Han, H.~Mao, and W.~J. Dally.
\newblock Trained ternary quantization.
\newblock In {\em ICLR}, 2017.

\bibitem{zhu2019simreal}
F.~Zhu, L.~Zhu, and Y.~Yang.
\newblock Sim-real joint reinforcement transfer for 3d indoor navigation.
\newblock In {\em Proceedings of the IEEE Conference on Computer Vision and
  Pattern Recognition (CVPR)}, 2019.

\bibitem{zhuang2018discrimination}
Z.~Zhuang, M.~Tan, B.~Zhuang, J.~Liu, Y.~Guo, Q.~Wu, J.~Huang, and J.~Zhu.
\newblock Discrimination-aware channel pruning for deep neural networks.
\newblock In {\em NIPS}, 2018.

\bibitem{zhuo2018scsp}
H.~Zhuo, X.~Qian, Y.~Fu, H.~Yang, and X.~Xue.
\newblock Scsp: Spectral clustering filter pruning with soft self-adaption
  manners.
\newblock {\em arXiv preprint arXiv:1806.05320}, 2018.

\end{thebibliography}
}

\end{document}